\documentclass[twoside]{article}

\usepackage[arxiv]{aistats2022}
\usepackage{booktabs}
\usepackage{xcolor}
\usepackage{subcaption}
\usepackage{graphicx}

\usepackage{calc}       
\newlength{\imagew}
\newlength{\imageh}
\newlength{\legendw}
\newlength{\legendh}
\newlength{\legendx}
\newlength{\legendy}
\newcommand{\graphicswithlegend}[6]{
  \setlength{\imagew}{#1}
  \settoheight{\imageh}{\includegraphics[width=\imagew]{#2}}

  \setlength{\legendw}{#3\imagew}
  \settoheight{\legendh}{\includegraphics[width=\legendw]{#4}}

  \setlength{\legendx}{\imagew}
  \addtolength{\legendx}{-\legendw}
  \addtolength{\legendx}{-#5\imagew}

  \setlength{\legendy}{\imageh}
  \addtolength{\legendy}{-\legendh}
  \addtolength{\legendy}{-#6\imageh}

  \includegraphics[width=\imagew]{#2}%
  \llap{
    \hspace{-\the\legendx}
    \raisebox{\legendy}{\includegraphics[width=\legendw]{#4}}
    \hspace{\the\legendx}
  }
}

\usepackage{mathtools}
\usepackage{amsthm}  
\usepackage{amsfonts}       

\DeclarePairedDelimiterX{\norm}[1]{\lVert}{\rVert}{#1}

\usepackage{adjustbox}
\usepackage{array}

\newcolumntype{R}[2]{%
    >{\adjustbox{angle=#1,lap=\width-(#2)}\bgroup}%
    l%
    <{\egroup}%
}
\newcommand{\mainfigsize}{0.32}

\newcommand{\otherfigsize}{0.48}

\usepackage{hyperref}       
\usepackage{url}            

\begin{document}

%

%

\twocolumn[

\aistatstitle{Identifying Layers Susceptible to Adversarial Attacks}

\aistatsauthor{ Shoaib Ahmed Siddiqui \And Thomas Breuel }

\aistatsaddress{ German Research Center for Artificial Intelligence (DFKI)\\
  TU Kaiserslautern\\
  \texttt{shoaib\_ahmed.siddiqui@dfki.de} \\ \And  NVIDIA Research \\
  \texttt{tbreuel@nvidia.com} } ]

\begin{abstract}
   In this paper, we investigate the use of pretraining with adversarial networks, with the objective of discovering the relationship between network depth and robustness.
   For this purpose, we selectively retrain different portions of VGG and ResNet architectures on CIFAR-10, Imagenette, and ImageNet using non-adversarial and adversarial data. Experimental results show that susceptibility to adversarial samples is associated with low-level feature extraction layers. Therefore, retraining of high-level layers is insufficient for achieving robustness. 
   Furthermore, adversarial attacks yield outputs from early layers that differ statistically from features for non-adversarial samples and do not permit consistent classification by subsequent layers.
   This supports common hypotheses regarding the association of robustness with the feature extractor, insufficiency of deeper layers in providing robustness, and large differences in adversarial and non-adversarial feature vectors.
\end{abstract}

\section{Introduction}

Deep neural networks often yield performance on test sets comparable to human performance~\cite{resnet}. However, at the same time, they have been found to be susceptible to imperceptible perturbations of inputs~\cite{szegedy2013intriguing,goodfellow2014explaining,madry2017towards,xie2019featuredenoising}.
These new samples crafted by an adversary with the aim of fooling the classifier are termed adversarial examples~\cite{szegedy2013intriguing}.
There has been a plethora of research in developing stronger defenses as well as stronger adversarial attacks to circumvent these defenses~\cite{goodfellow2014explaining,madry2017towards,xie2019featuredenoising,zhang2019trades,wong2020fast,akhtar2018defense,naseer2020selfsup,folz2020robustness_s2s,li2020enhancing}.
However, the reasons for their existence are still poorly understood~\cite{goodfellow2014explaining,ilyas2019adversarialexamplesarenotbugs,wang2019highfreq}.
Understanding these differences between deep neural networks and human perception is important both in order to understand the mathematical and statistical structure of such networks, as well as to protect systems against attacks.

Deep neural networks automate the task of feature extraction, obviating the need for hand-engineering features.
Such networks are thought of as consisting of initial feature extraction layers and high-level layers responsible for learning decision boundaries.
In fact, in many cases, initial feature extraction layers are often reused in practice between different datasets and tasks to speed up convergence (commonly known as transfer learning~\cite{zhuang2020transfer}).
In the context of adversarial samples, if we could reuse feature extraction layers, it would greatly speed up research in adversarial samples, since adversarial samples could be studied on pre-extracted data.
If susceptibility to adversarial samples is associated with high-level layers, it would also give us insights into the nature of adversarial phenomena and suggest that adversarial samples might be related primarily to the formation of decision boundaries by the high-level layers.
This idea has been leveraged in techniques such as large-margin classification to achieve adversarial robustness~\cite{elsayed2018large}.

\begin{figure*}[t]
    \centering
    \includegraphics[width=1.0\textwidth]{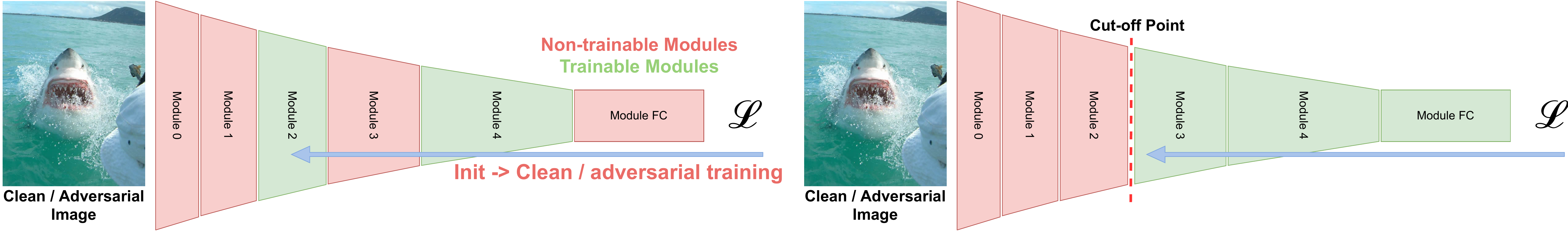}
    \caption{Overview of the training methodology where a set of of blocks is reinitialized and retrained (using either clean or adversarial samples) while others are kept frozen after loading the pretrained model. One special case is that of a single cut-off point (right) where the network from the beginning to the cut-off point or from the cut-off point to the end is trained while keeping the other part fixed.}
    \label{fig:overview}
\end{figure*}

To test these ideas,  we use a novel block-wise retraining protocol.
In particular, considering a pretrained model, either trained non-adversarially or adversarially, we reinitialize and retrain a particular set of blocks either conventionally for an adversarially pretrained model or adversarially for a conventionally pretrained model.
The key findings of this paper can be summarized as follows:
\begin{itemize}
    \item Adversarial retraining of just low-level/early layers is associated with strong reductions in susceptibility to adversarial samples.
    \item Adversarial retraining of just high-level/late layers fails to result in robustness to adversarial samples.
    \item The distributions of feature vectors from non-adversarial and adversarial inputs differ substantially at all levels; therefore, susceptibility to adversarial attacks is associated with the early generation of feature vectors that do not occur in non-adversarial images.
    \item Adversarial training results in weights for early layers that bring the distribution of feature vectors for adversarial samples back to the distribution of feature vectors for non-adversarial samples.
\end{itemize}
Overall, these results show that susceptibility to adversarial samples is primarily a phenomenon associated with early layers and low-level feature extraction.
Adversarial samples do not merely transform the appearance of one image into that of another, but instead generate novel feature vectors that result in novel activation patterns in late layers and high-level, class-specific feature vectors.

\section{Related Work}

\begin{figure*}[t]
\centering
    
    \begin{subfigure}[b]{0.9\textwidth}
    \centering
    \begin{subfigure}[b]{\mainfigsize\textwidth}
        \graphicswithlegend{\columnwidth}{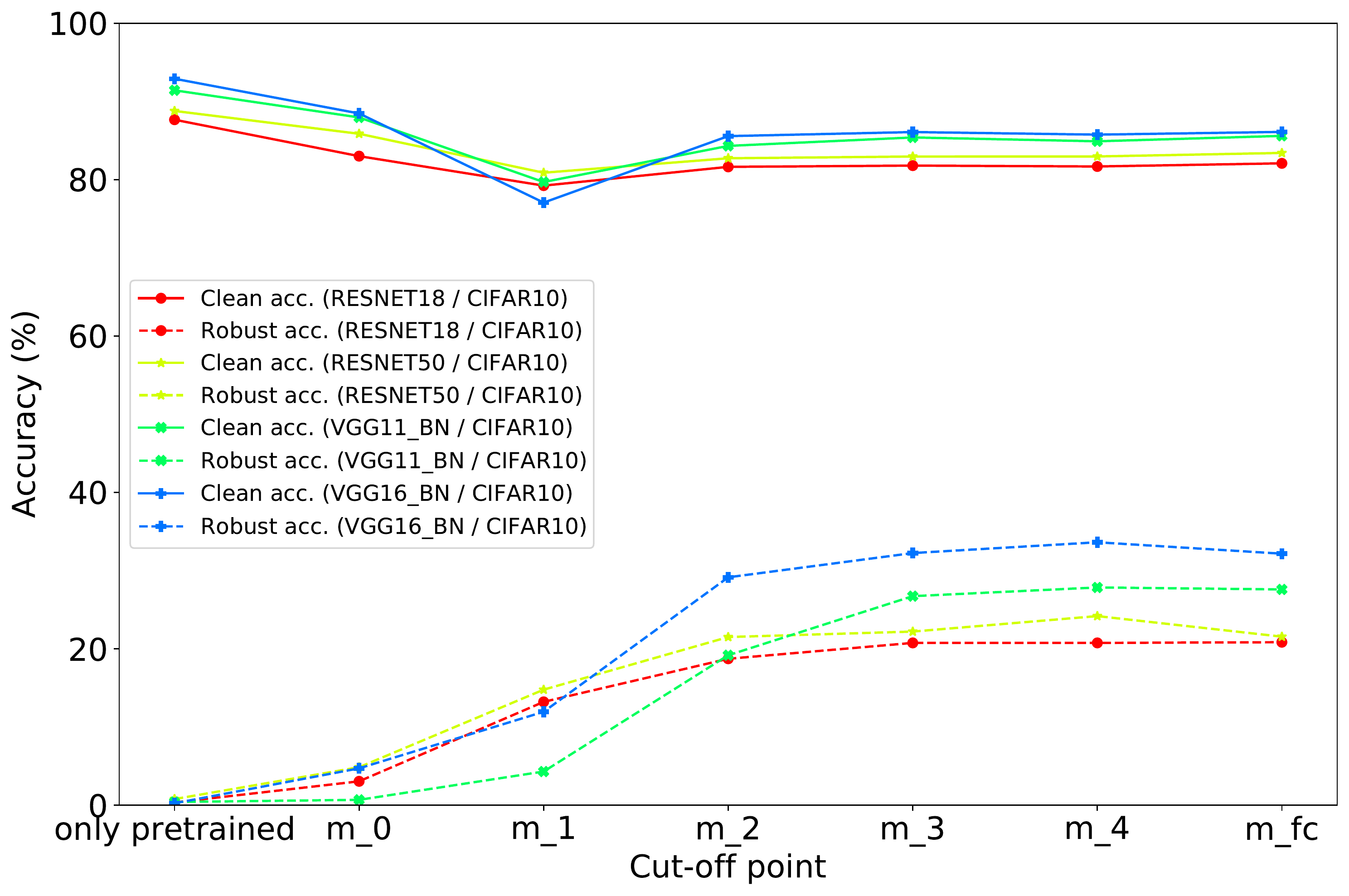}
                      {0.1}{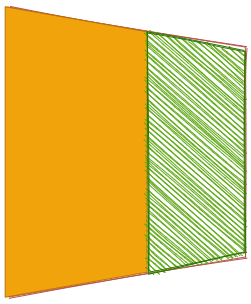}{0.88}{0.7}
        \vspace{-4mm}
        \caption{Adv. retraining up to cut-off}
    \end{subfigure}
    ~
    \begin{subfigure}[b]{\mainfigsize\textwidth}
        \graphicswithlegend{\columnwidth}{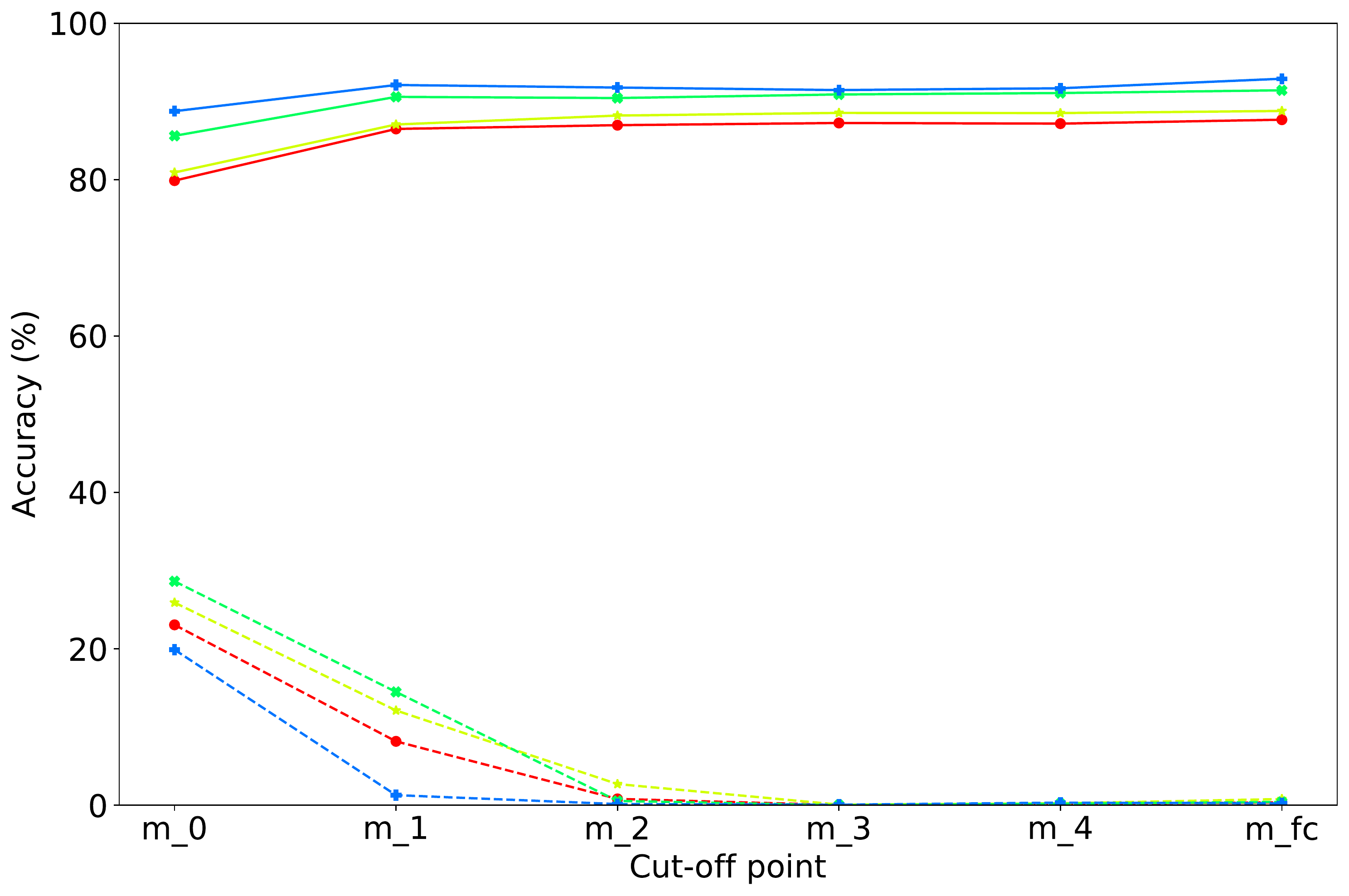}
                      {0.1}{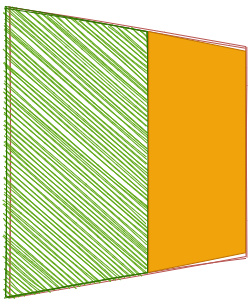}{0.88}{0.7}
        \vspace{-4mm}
        \caption{Adv. retraining after cut-off}
    \end{subfigure}
    
    \begin{subfigure}[b]{\mainfigsize\textwidth}
        \graphicswithlegend{\columnwidth}{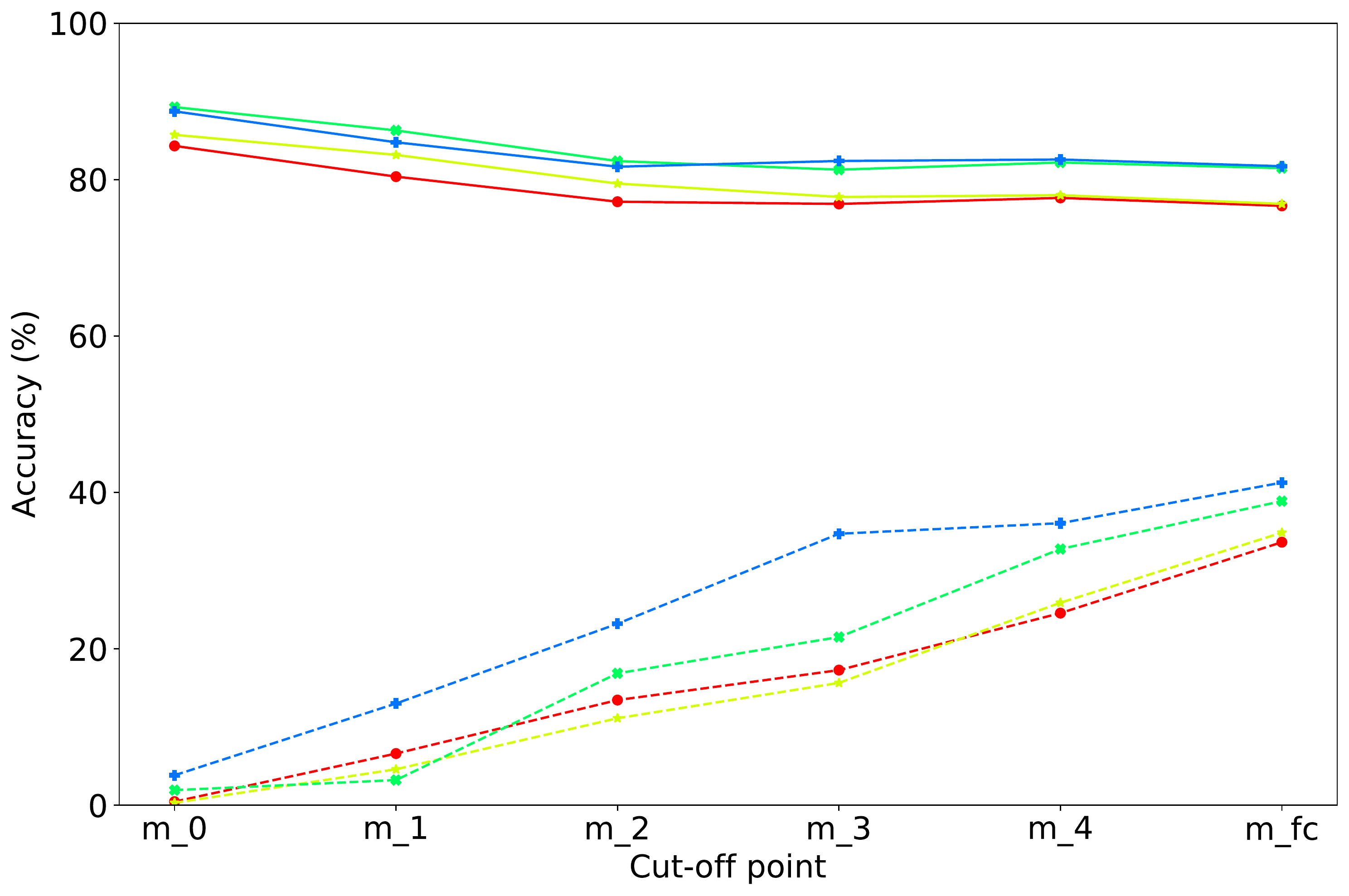}
                      {0.1}{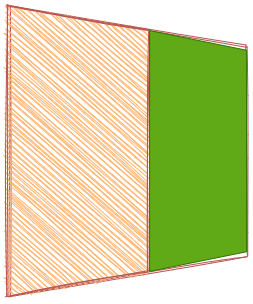}{0.88}{0.7}
        \vspace{-4mm}
        \caption{Con. retraining after cut-off}
    \end{subfigure}
    ~
    \begin{subfigure}[b]{\mainfigsize\textwidth}
        \graphicswithlegend{\columnwidth}{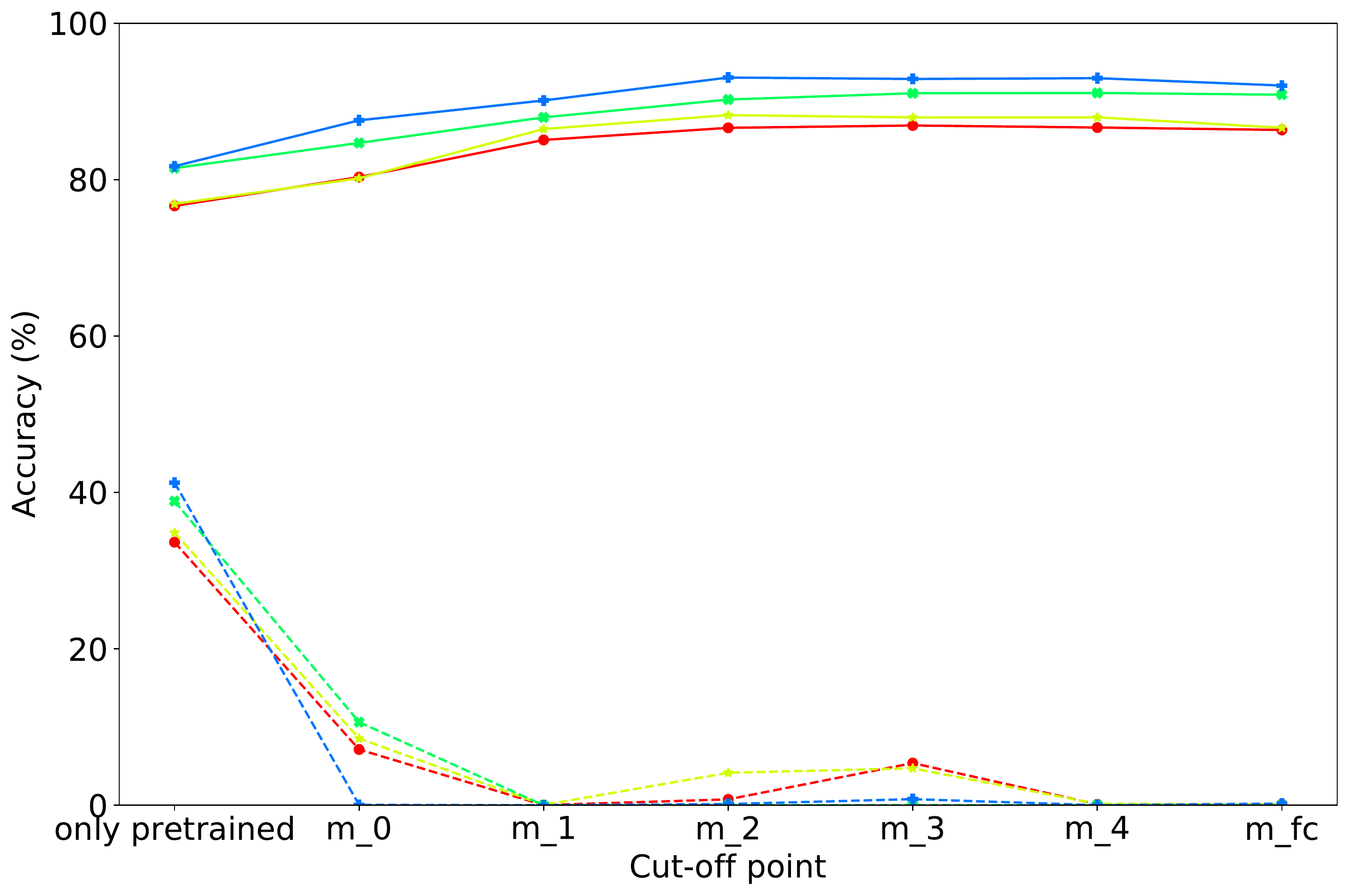}
                      {0.1}{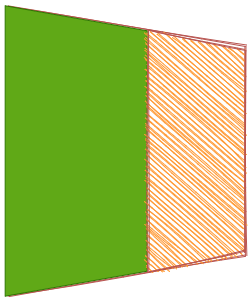}{0.88}{0.7}
        \vspace{-4mm}
        \caption{Con. retraining up to cut-off}
    \end{subfigure}
    \end{subfigure}
    \hspace{-25mm}
    \begin{subfigure}[b]{0.09\textwidth}
    \raisebox{0.5\height}{\includegraphics[width=\textwidth]{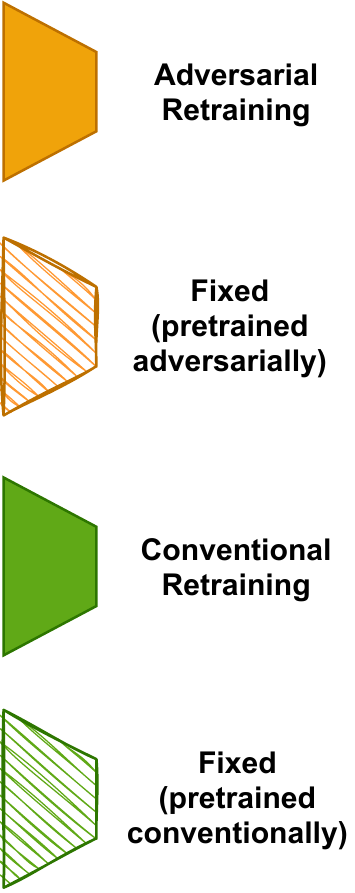}}
    \end{subfigure}
    
    \caption{Partial retraining of VGG and ResNet architectures on CIFAR-10 shows that robustness to adversarial samples is achieved if and only if the weights for early layers were either pretrained or retrained with adversarial samples.
    Retraining up to \textit{m\_fc} refers to full retraining of the pretrained model. Similarly, retraining after \textit{m\_fc} refers to the pretrained model without any partial retraining. In cases where the performance of the pretrained model is not naturally incorporated in the plot, \textit{only pretrained} refers to the performance of the original pretrained network without any partial retraining.}
    \label{fig:cifar10_results}
\end{figure*}

Susceptibility to adversarial samples~\cite{goodfellow2014explaining,madry2017towards,xie2019featuredenoising,zhang2019trades,wong2020fast,akhtar2018defense,naseer2020selfsup,folz2020robustness_s2s,li2020enhancing} has been explained
in terms of overreliance on texture for classification~\cite{geirhos2018imagenettrainedshapebias}, excessive invariance~\cite{jacobsen2018excessive}, over-reliance on high-frequencies~\cite{wang2019highfreq}, piece-wise linear nature of deep networks~\cite{goodfellow2014explaining}, or even a bias present in the dataset itself~\cite{ilyas2019adversarialexamplesarenotbugs}.

One of the most effective defenses against adversarial samples has been {\bf robust optimization}~\cite{goodfellow2014explaining,madry2017towards,xie2019featuredenoising,zhang2019trades,wong2020fast}, where the model is trained on adversarial samples rather than clean samples.
Adversarial training has been combined with transfer learning~\cite{shafahi2019transferablerobustness,jeddi2020simple}, using the typical approach of reusing the low-level (``feature extraction'') layers; our work extends this approach to systematically determine which portions of networks can be retrained to achieve adversarial robustness.
{\bf Image-denoising-based defenses}~\cite{akhtar2018defense,naseer2020selfsup,folz2020robustness_s2s,li2020enhancing} attempt to remove the noise introduced by adversarial attacks; the success of such approaches suggests that low-level layers in networks may be important for adversarial defenses but do not exclude the possibility that high-level layers may be important as well.
Recent results indicate that vision transformers may be more robust to adversarial attacks than convolutional architectures~\cite{shao2021vitrobust} when trained adversarially.

Work on {\bf adversarial example detection}~\cite{roth2019odds} attempts to identify adversarial examples based on distinctive feature vectors.
Other work~\cite{mao2019metric,li2020towards,bai2021improving,bai2021improving} also compares the activations in the penultimate layer of networks between adversarial and non-adversarial samples. In this paper, we compare and analyze activations from adversarial and non-adversarial samples at different depths and for different training modalities, yielding new insights into the origin of adversarial samples.

\section{Methods}

\begin{figure*}[t]
\centering
    
    \begin{subfigure}[b]{0.9\textwidth}
    \centering
    \begin{subfigure}[b]{\mainfigsize\textwidth}
        \graphicswithlegend{\columnwidth}{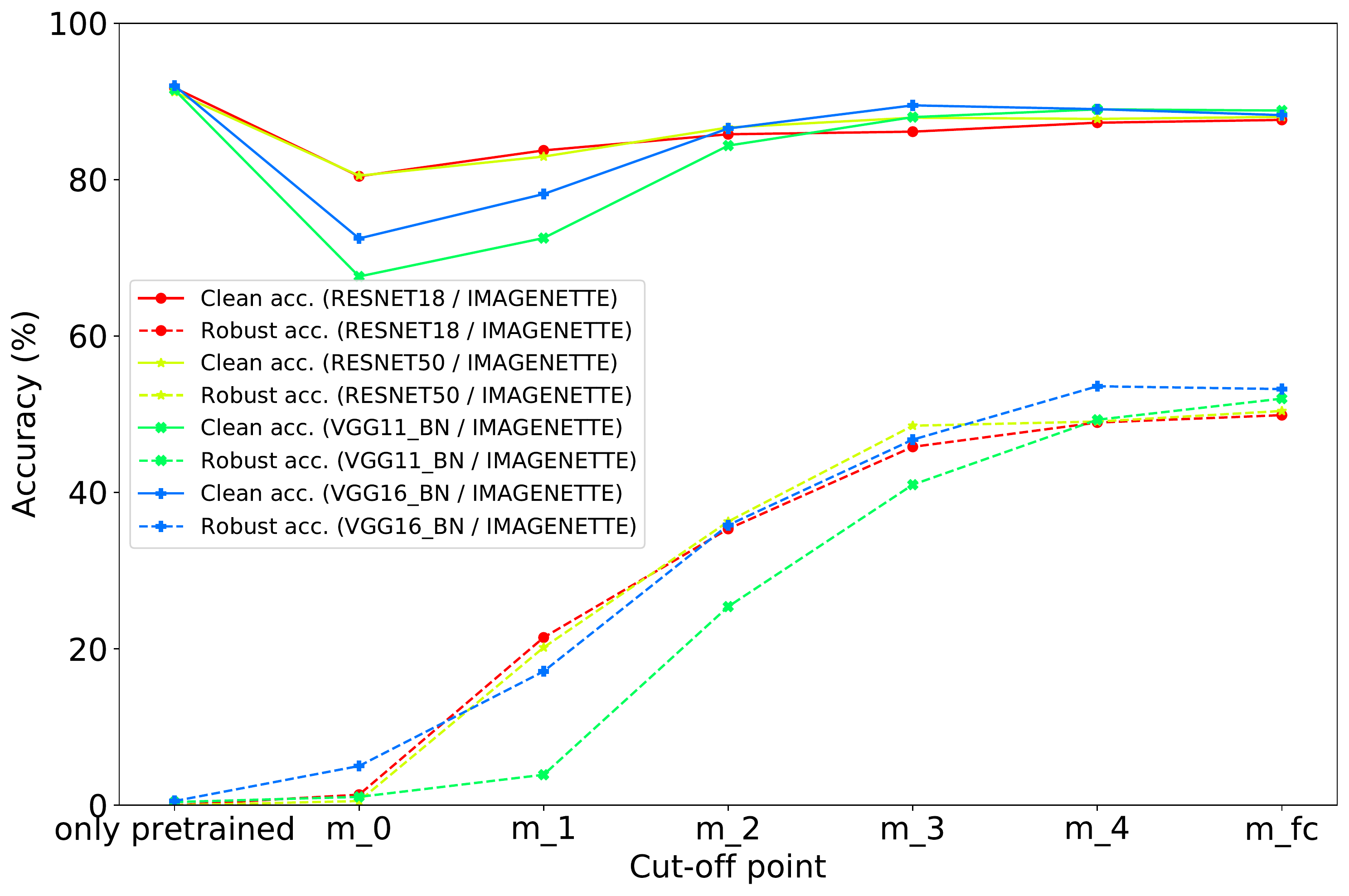}
                      {0.1}{Figures/illustrations/training_type_adv_upto.png}{0.88}{0.7}
        \vspace{-4mm}
        \caption{Adv. retraining up to cut-off}
    \end{subfigure}
    ~
    \begin{subfigure}[b]{\mainfigsize\textwidth}
        \graphicswithlegend{\columnwidth}{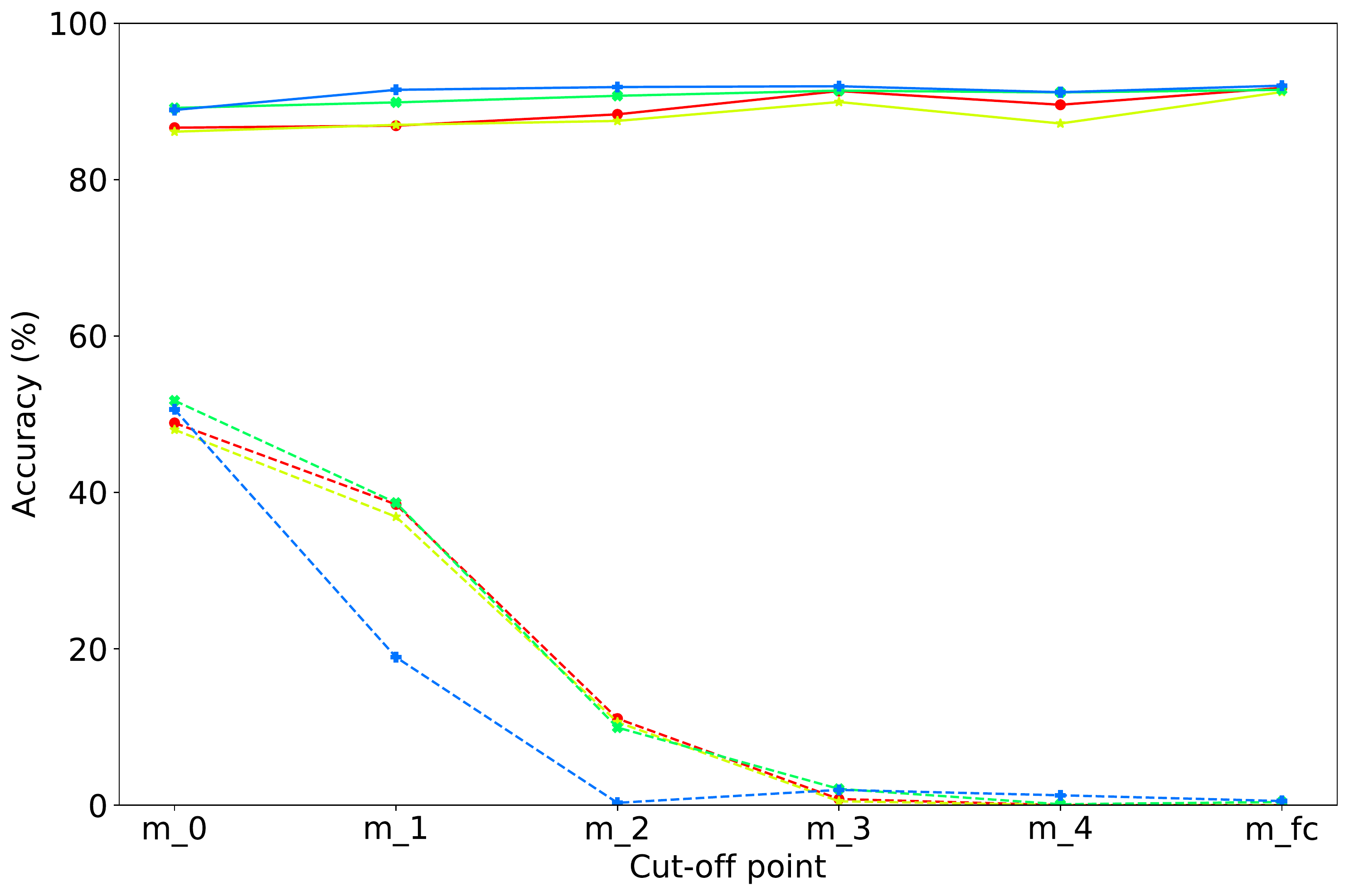}
                      {0.1}{Figures/illustrations/training_type_adv_after.png}{0.88}{0.7}
        \vspace{-4mm}
        \caption{Adv. retraining after cut-off}
    \end{subfigure}
    
    \begin{subfigure}[b]{\mainfigsize\textwidth}
        \graphicswithlegend{\columnwidth}{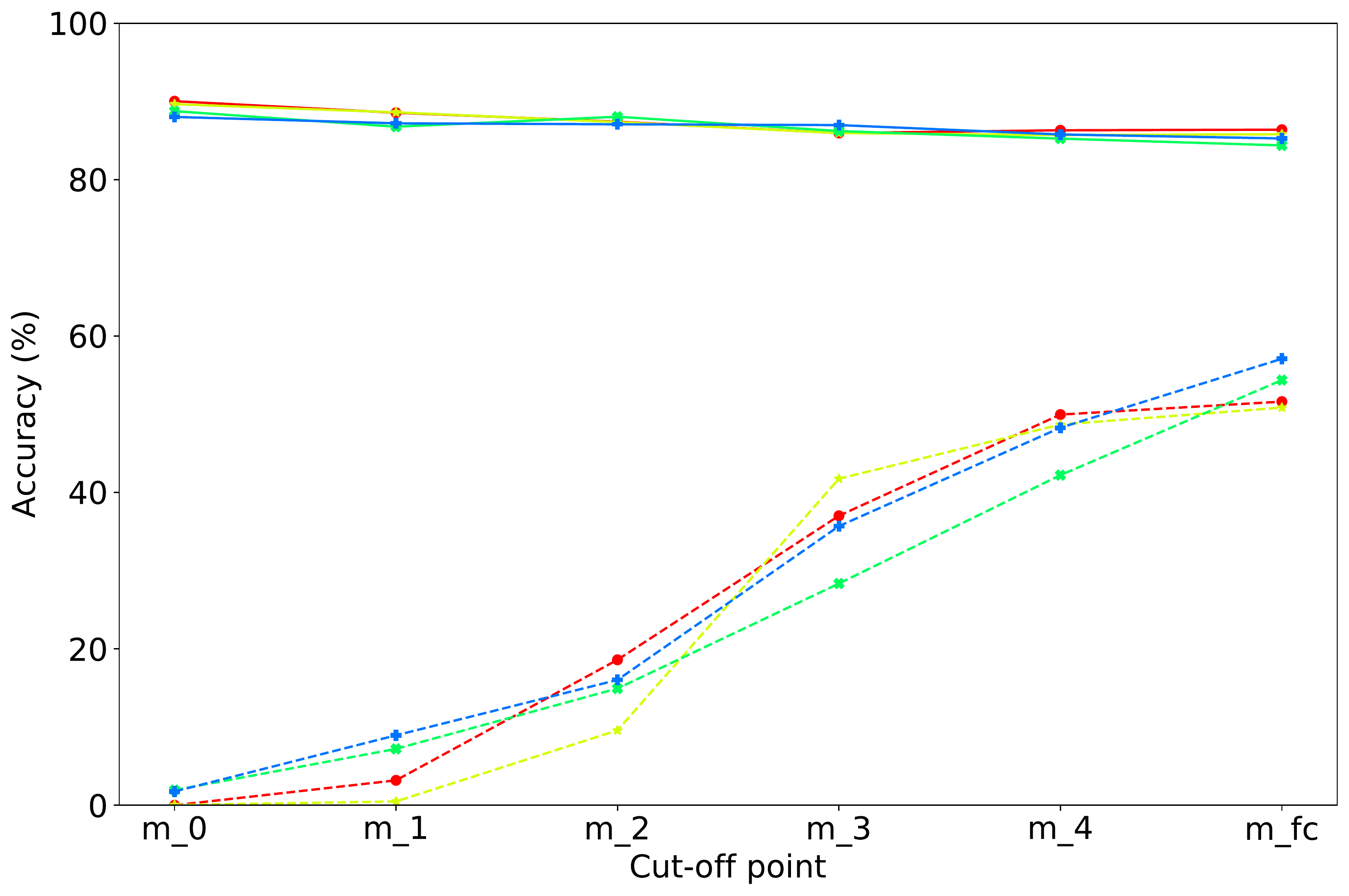}
                      {0.1}{Figures/illustrations/training_type_clean_after.png}{0.88}{0.7}
        \vspace{-4mm}
        \caption{Con. retraining after cut-off}
    \end{subfigure}
    ~
    \begin{subfigure}[b]{\mainfigsize\textwidth}
        \graphicswithlegend{\columnwidth}{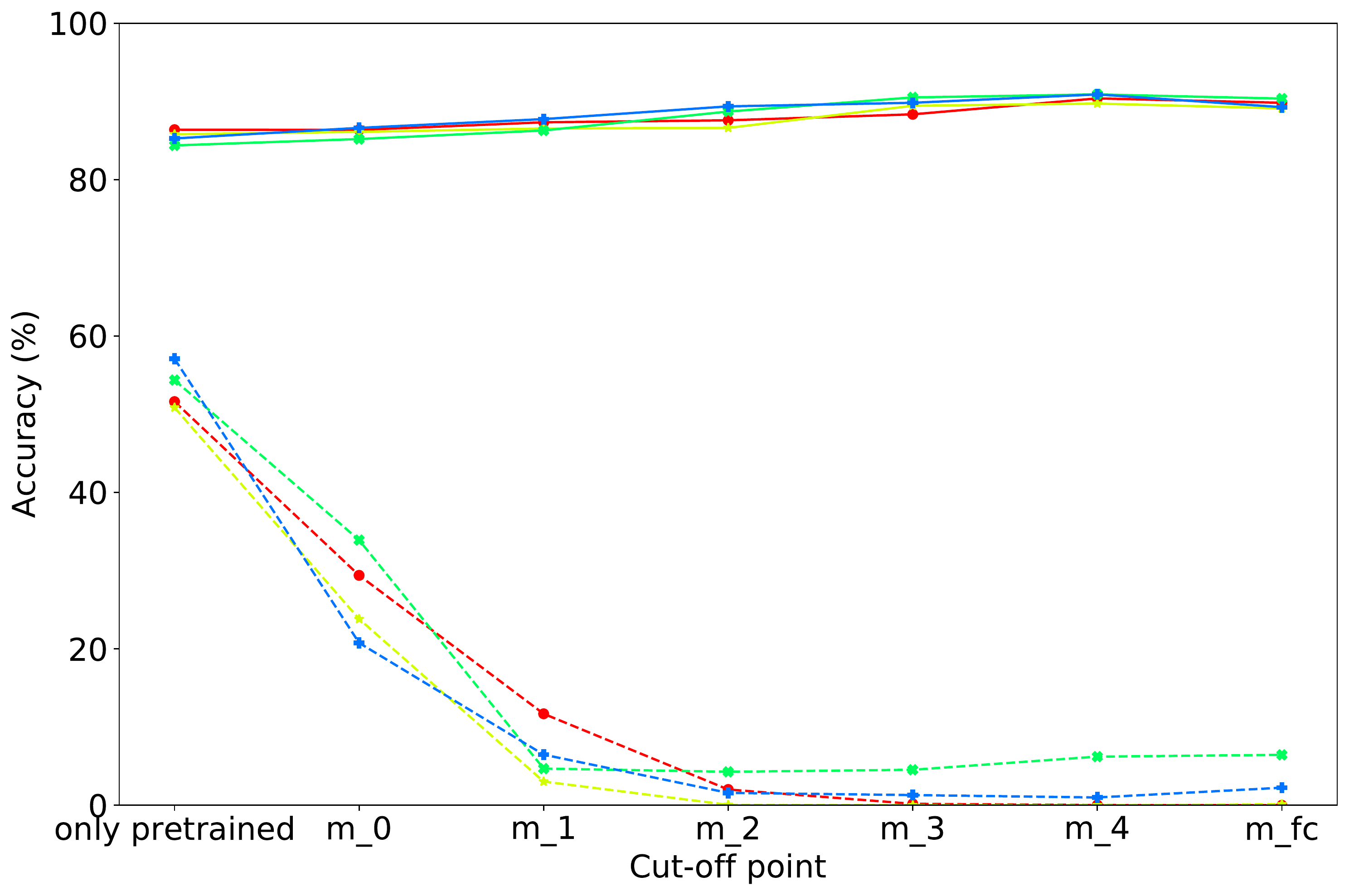}
                      {0.1}{Figures/illustrations/training_type_clean_upto.png}{0.88}{0.7}
        \vspace{-4mm}
        \caption{Con. retraining up to cut-off}
    \end{subfigure}
    \end{subfigure}
    \hspace{-25mm}
    \begin{subfigure}[b]{0.09\textwidth}
    \raisebox{0.5\height}{\includegraphics[width=\textwidth]{Figures/illustrations/training_type_legend_vertical.pdf}}
    \end{subfigure}
    
    \caption{Results on Imagenette are primarily consistent with all the findings on the CIFAR-10 dataset despite a larger input size of $224 \times 224$. However, we see that since the size of the objects is larger in this case, the importance of the higher-level modules increases.}
    \label{fig:imagenette_results}
\end{figure*}

Layers present in a network are often thought of as operating at different semantic levels, where initial layers respond to basic features such as edges or gradients, while higher layers represents complete objects or some prominent parts of it~\cite{yosinski2015DeepVisToolbox}. 
In order to analyze the role of different layers of the network in terms of their susceptibility to adversarial noise, we use a block-wise retraining protocol. Given a model (ResNet or VGG in our case), we split the model into different modules. Both ResNet and VGG models are naturally dissected into six modules by the down-sampling layers within the network (max-pooling in the case of VGG and convolutional layer with a stride of 2 in the case of ResNets).
An overview of the method is presented in Fig.~\ref{fig:overview}.

We first pretrain the complete network either conventionally or adversarially.
Conventional training refers to training on clean images while adversarially training refers to training on adversarial images computed using a particular attack method.
We follow the adversarial training recipe from Madry et al. (2017)~\cite{madry2017towards} where we train the model on adversarial images computed using Projected-Gradient Descent (PGD) attack.
Once the model is pretrained, we reinitialize and retrain a set of modules of the network adversarially for the conventionally pretrained model or conventionally for the adversarially pretrained model, while keeping the weights for the rest of the modules fixed.
Our main experiments rely on a single splitting point for the network, where we only retrain all the layers before or after the cut-off point.

\subsection{Datasets}

We validate our findings by testing on three different datasets including CIFAR-10~\cite{krizhevsky2014cifar}, ImageNet~\cite{ILSVRC15} and Imagenette~\cite{imagenette}.
CIFAR-10~\cite{krizhevsky2014cifar} is a 10 class dataset comprising of low-resolution images ($32 \times 32$) with 50000 training and 10000 test samples.
We also include the large-scale high-resolution ImageNet dataset with 1.28M training and 50000 validation samples to evaluate our hypothesis\footnote{The validation set serves the purpose of the test set in our case as direct access to the test set is not available. Therefore, no hyperparameters are tuned directly on the validation set.}.
Finally, we include a small subset of ImageNet called Imagenette with only 10 classes but high-resolution images (9469 training and 3925 test samples).

\subsection{Experimental Protocol}

\begin{figure*}[t]
\centering
    
    \begin{subfigure}[b]{0.9\textwidth}
    \centering
    \begin{subfigure}[b]{\mainfigsize\textwidth}
        \graphicswithlegend{\columnwidth}{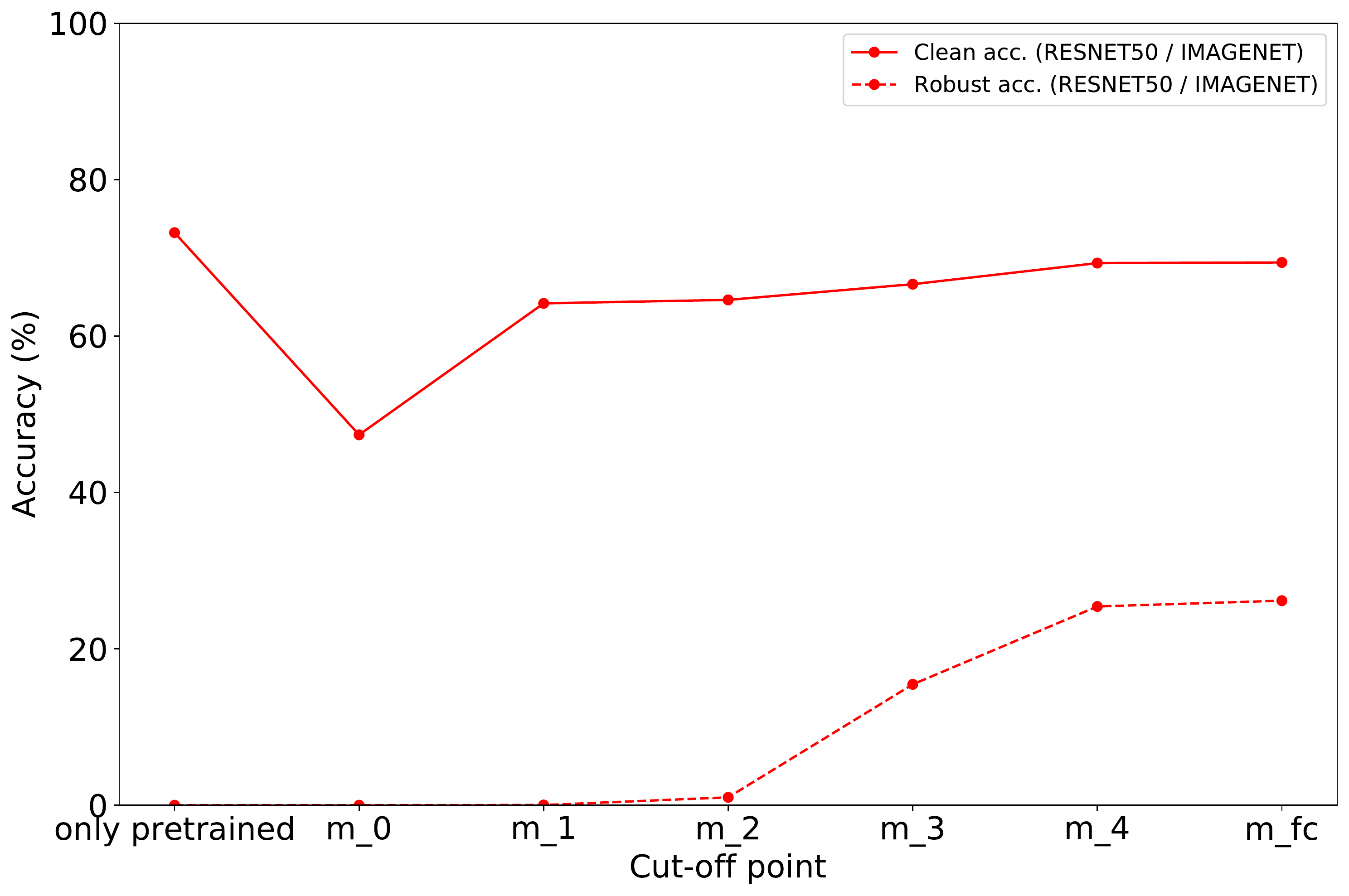}
                      {0.1}{Figures/illustrations/training_type_adv_upto.png}{0.88}{0.7}
        \vspace{-4mm}
        \caption{Adv. retraining up to cut-off}
    \end{subfigure}
    ~
    \begin{subfigure}[b]{\mainfigsize\textwidth}
        \graphicswithlegend{\columnwidth}{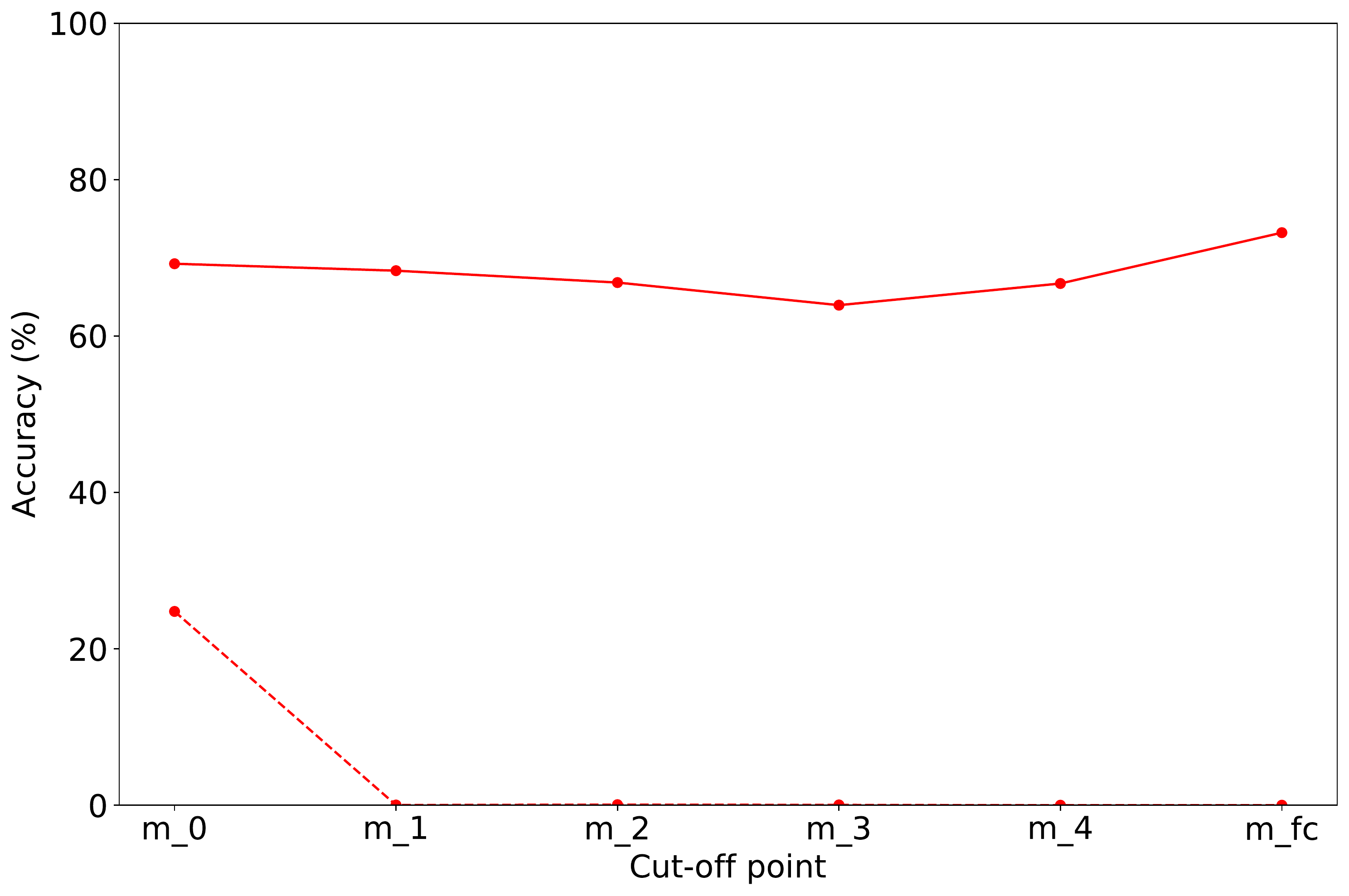}
                      {0.1}{Figures/illustrations/training_type_adv_after.png}{0.88}{0.7}
        \vspace{-4mm}
        \caption{Adv. retraining after cut-off}
    \end{subfigure}
    
    \begin{subfigure}[b]{\mainfigsize\textwidth}
        \graphicswithlegend{\columnwidth}{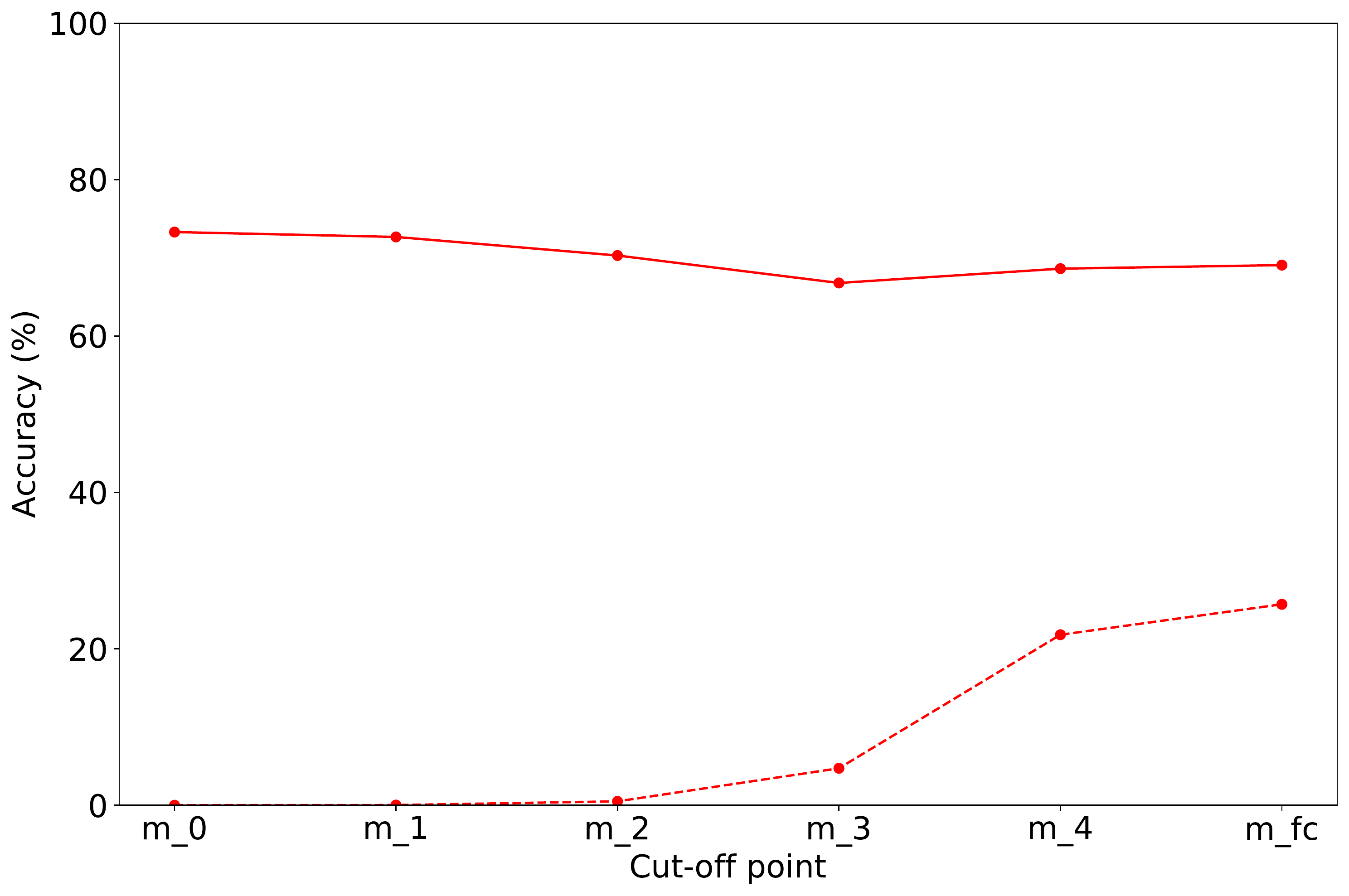}
                      {0.1}{Figures/illustrations/training_type_clean_after.png}{0.88}{0.7}
        \vspace{-4mm}
        \caption{Con. retraining after cut-off}
    \end{subfigure}
    ~
    \begin{subfigure}[b]{\mainfigsize\textwidth}
        \graphicswithlegend{\columnwidth}{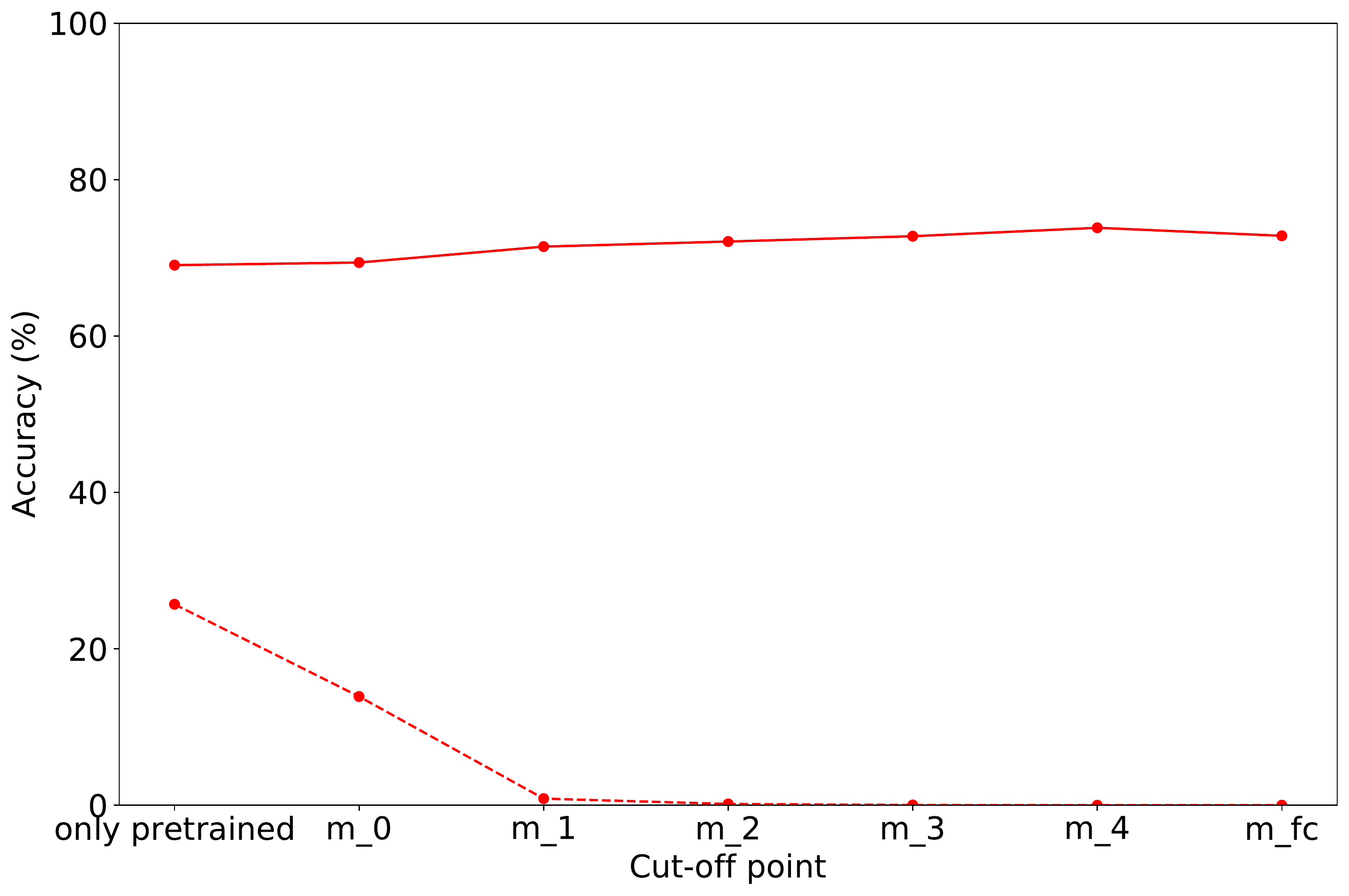}
                      {0.1}{Figures/illustrations/training_type_clean_upto.png}{0.88}{0.7}
        \vspace{-4mm}
        \caption{Con. retraining up to cut-off}
    \end{subfigure}
    \end{subfigure}
    \hspace{-25mm}
    \begin{subfigure}[b]{0.09\textwidth}
    \raisebox{0.5\height}{\includegraphics[width=\textwidth]{Figures/illustrations/training_type_legend_vertical.pdf}}
    \end{subfigure}
    
    \caption{ImageNet results with ResNet-50 architecture are consistent with the results on Imagenette or CIFAR-10, but we observe a shift in the cut-off point resulting in a rise in the importance of the higher-level modules. This can be partly explained by the larger size of ImageNet where initial modules themselves are insufficient to provide robustness since most of the parameters lie in the higher-level modules.}
    \label{fig:imagenet_results}
\end{figure*}

All the CIFAR models were trained on a single GPU (NVIDIA RTXA6000) with Adam optimizer with an initial learning rate of 0.001 and a batch size of 128. The models were trained for 300 epochs, with a cosine decay in the learning rate after every epoch.
All our results are based on $L_{\infty}$ norm-based attacks.
For adversarial training, we used an epsilon of 8/255 and epsilon per iteration of 2/255. We used PGD-based adversarial training~\cite{madry2017towards,xie2019featuredenoising} where the number of iterations was fixed to 7.
We use identical training settings for Imagenette.

For the ImageNet dataset, we use fast adversarial training~\cite{wong2020fast} to speed up the training process. 
Fast adversarial training uses a random start followed by a single step in the direction of the gradient which makes it equivalent to PGD-1. We use an epsilon of 4/255 for ImageNet as per the common practice~\cite{wong2020fast,xie2020smooth}. The model was trained using 8 GPUs (NVIDIA RTXA6000) with synchronized batch-norm using SGD with an initial learning rate of 0.256, a momentum of 0.875, and a batch size of 256, where the learning rate was reduced by a factor of 10 after the $30^{th}, 60^{th}$ and the $90^{th}$ epoch.
We used a weight decay of 0.0001 to train all our models.

For both Imagenette and CIFAR-10, we add clean samples to the batch during adversarial retraining (we use a ratio of 50:50 for clean and adversarial samples in the batch).
This inclusion of clean examples helps maintain the clean accuracy of the model.
However, we do not include any clean samples when retraining ResNet-50 on ImageNet.

For model evaluation, we use PGD-200~\cite{xie2019featuredenoising,xie2020smooth,madry2017towards} with a single restart. 
Stronger attacks do exist~\cite{croce2020autoattack}, but the objective of this paper is to determine the relative susceptibility of layers.
Therefore, absolute numbers are not particularly important in our case.
It is important to mention that we attack the model using the actual target during evaluation rather than the prediction.
This ensures that the robust accuracy does not accidentally inflate due to the attack moving the class from wrong to the correct label.
However, we still attack the prediction of the model during adversarial training to avoid the \textit{label-leaking} effect~\cite{kurakin2016labelleaking} as per the common adversarial training recipe.

\subsection{Model Architectures}

We evaluated the VGG~\cite{vgg} and ResNet~\cite{resnet} model families imported from the TorchVision~\cite{torchvision} model repository. We specifically evaluated VGG-11 and VGG-16, where both of these models were equipped with batch-norm. In order for these architectures to work on CIFAR, we replace the average pooling layer before the classification head (which outputs a $7 \times 7$ tensor) with Global Average Pooling (GAP) layer, which reduces the dimensionality to $1 \times 1$. This is similar to the residual architecture~\cite{resnet}. 
We include ResNet-18 and ResNet-50 within the residual family~\cite{resnet} for our experiments with identical architecture across datasets.

\section{Key Results}

We evaluate the block-wise susceptibility of four models (ResNet-18, ResNet-50, VGG-11, VGG-16) belonging to two major model families (ResNet~\cite{resnet} and VGG~\cite{vgg}) on three image recognition datasets (CIFAR-10~\cite{krizhevsky2014cifar}, Imagenette~\cite{imagenette} and ImageNet~\cite{ILSVRC15}) using our block-wise retraining protocol.
The primary results are divided into four different training settings that we evaluate which include (i) adversarial retraining before the cut-off, (ii) adversarial retraining after the cut-off, (iii) conventional retraining before the cut-off, and (iv) conventional retraining after the cut-off.

\subsection{CIFAR-10}

Our results on CIFAR-10 are visualized in Fig.~\ref{fig:cifar10_results}.
In all four different settings, it is evident that adversarial performance improves with adversarial training of early layers.
This also indicates that the main discrepancy between conventionally and adversarially trained models lies at the lower-level features.
For conventional retraining, just retraining the initial two modules evades the robustness of the network, while having a marginal effect on clean accuracy. 
This highlights the fact that the initial modules are the most distinct ones between conventionally and adversarially trained models.

\subsection{Imagenette}

We see a similar trend on Imagenette in terms of retraining as CIFAR-10 where adversarial retraining of initial modules is important for obtaining robustness (Fig.~\ref{fig:imagenette_results}).
However, we see a relative increase in the importance of higher-level modules as compared to CIFAR-10.
This can be attributed to the larger image size, where the object occupies a larger fraction of the image, requiring a larger effective receptive field of the network.

\subsection{ImageNet}

The results on ImageNet are again consistent with our prior results on CIFAR-10 and Imagenette.
But the point is shifted where mid-level modules (m\_2, m\_3, and m\_4) also play a dominant role for robustness as evident from Fig.~\ref{fig:imagenet_results}. 
Our evaluation is limited to ResNet-50 trained using fast adversarial training~\cite{wong2020fast} on ImageNet~\cite{ILSVRC15}.
Looking at the performance in the case where the model is retrained after the cut-off point, we see the same trend where excluding the first two modules results in poor robustness of the model.

\section{Analysis}

Based on our preliminary findings, we analyze particular aspects of the models in more detail.
This includes extending our analysis to every layer, evaluating all possible combinations of modules as well as evaluating layer robustness to reinitialization.

\subsection{Per-Layer Analysis}

\begin{figure}[t]
\centering
    \includegraphics[width=0.48\textwidth]{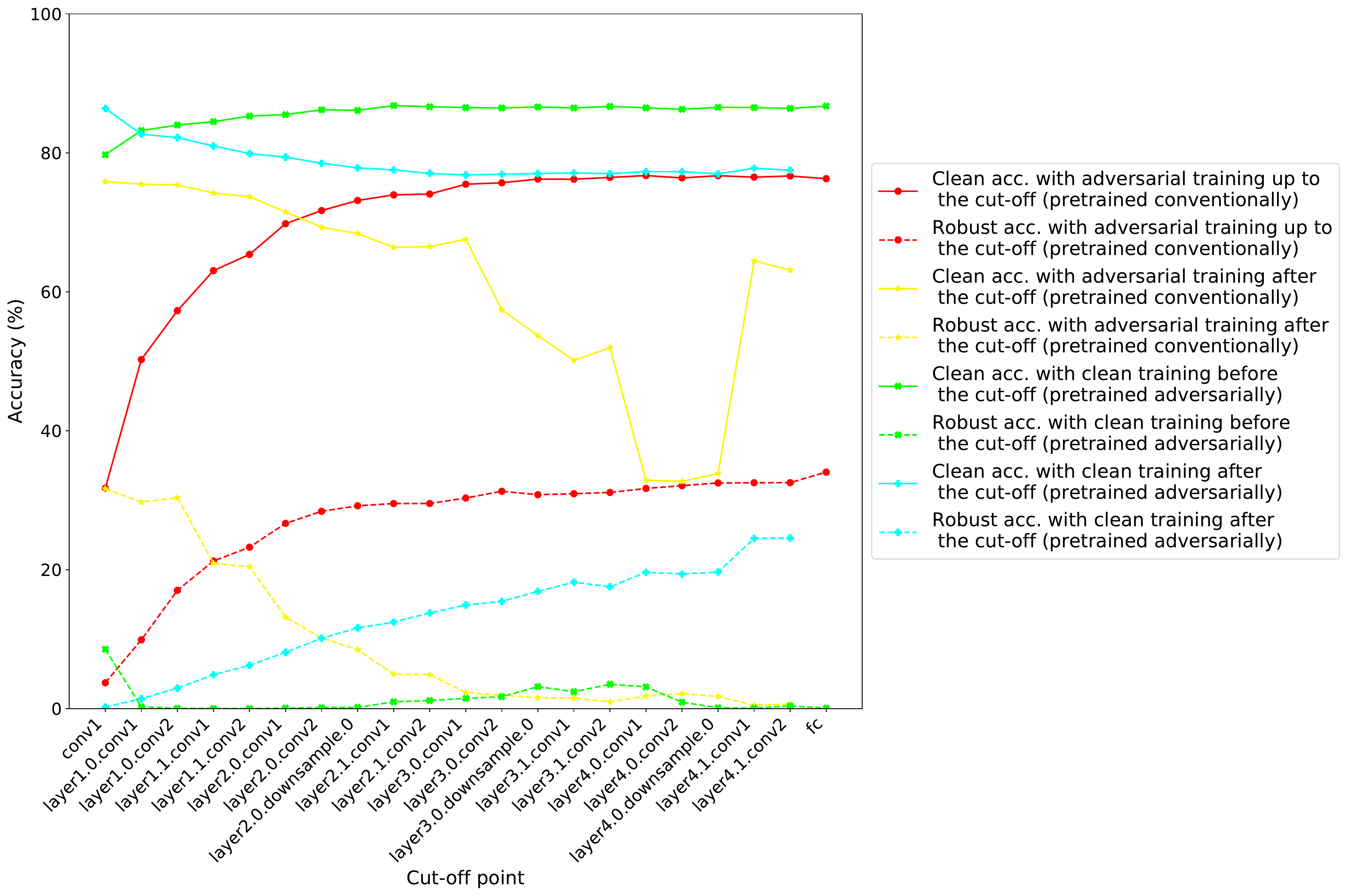}
    \caption{Results decomposed at the level of layers rather than modules for the ResNet-18 architecture on CIFAR-10. These results are consistent with the module-based results as the initial layers are part of the initial modules of the network. We do not observe any significant contribution from a single layer, but rather, the contribution from different layers accumulates.}
    \label{fig:cifar10_resnet18_layer_results}
\end{figure}

We perform a more fine-grained layer-wise retraining evaluation for ResNet-18 on CIFAR-10 where we move away from complete modules which comprise of a different number of layers at each level and focus on the individual layers themselves. 
This analysis decouples the aggregation artifacts and highlights if there are other layers that are equally important as the initial layers in the network. 
The results for the per-layer cut-off experiment are visualized in Fig.~\ref{fig:cifar10_resnet18_layer_results}.
The analysis show that the gains are flattened out after the inclusion of the first seven layers, which is consistent with the results from the module-based experiment as these layers form the initial modules of the network. We observe the most significant gain for the first layer in the network, which is referred to as \textit{m\_0} in our previous results.

It is interesting to note that there is a sudden drop in clean accuracy in Fig.~\ref{fig:cifar10_resnet18_layer_results} when retraining module 4 adversarially. This is an interesting phenomenon that is likely related to the fact that for small input images like those found in CIFAR-10, module 4 already performs a kind of \textit{global classification}. The effect does not seem to be related to adversarial robustness. However, this observation highlights that module/layer-wise retraining is a technique that allows us to discover and analyze other unexpected effects in deep neural networks.

\subsection{Module Combinations}

\begin{figure}[t]
\centering
    \begin{subfigure}[b]{0.23\textwidth}
        \includegraphics[width=\textwidth]{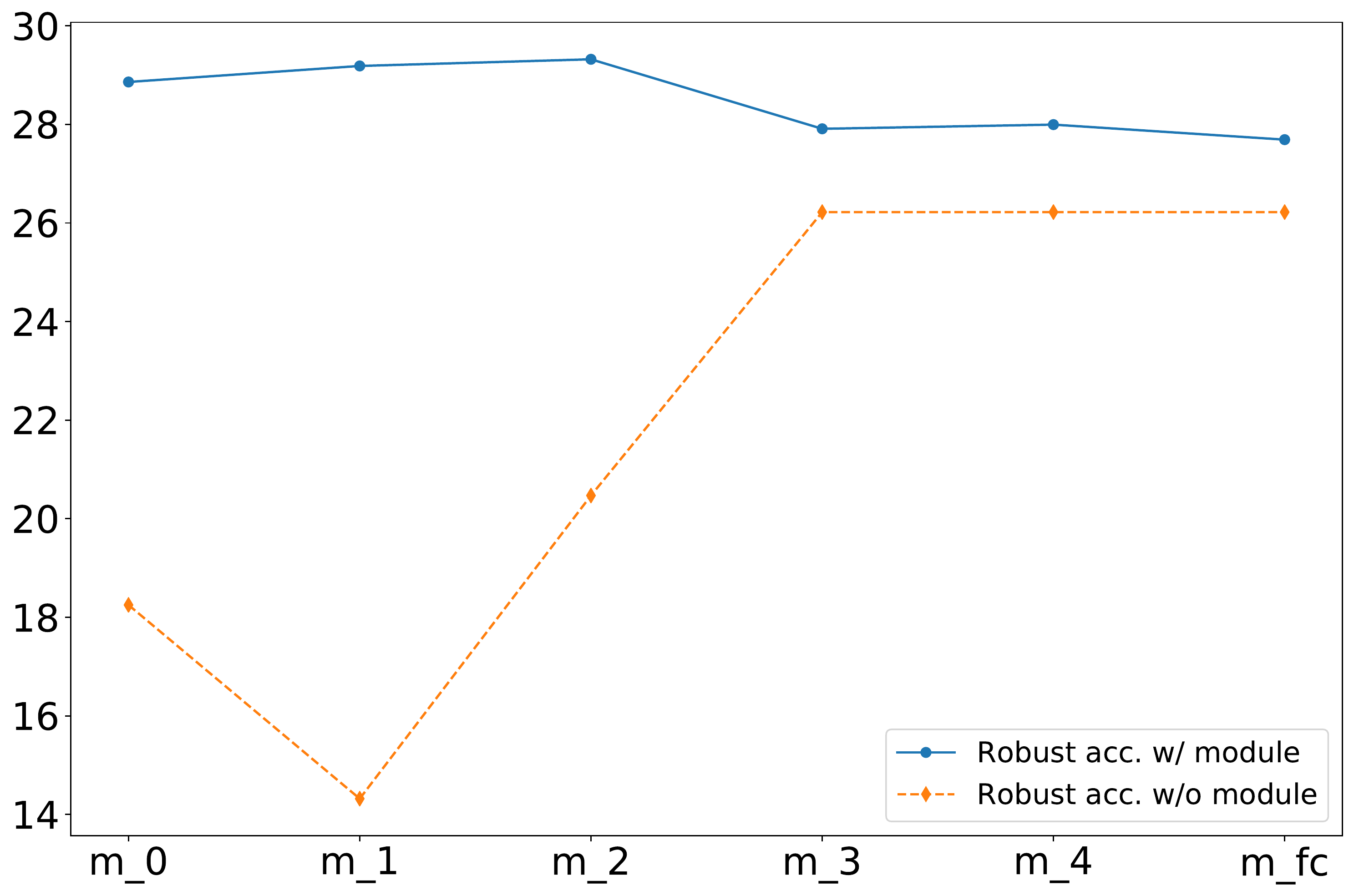}
        \caption{Adversarial retraining}
    \end{subfigure}
    ~
    \begin{subfigure}[b]{0.23\textwidth}
        \includegraphics[width=\textwidth]{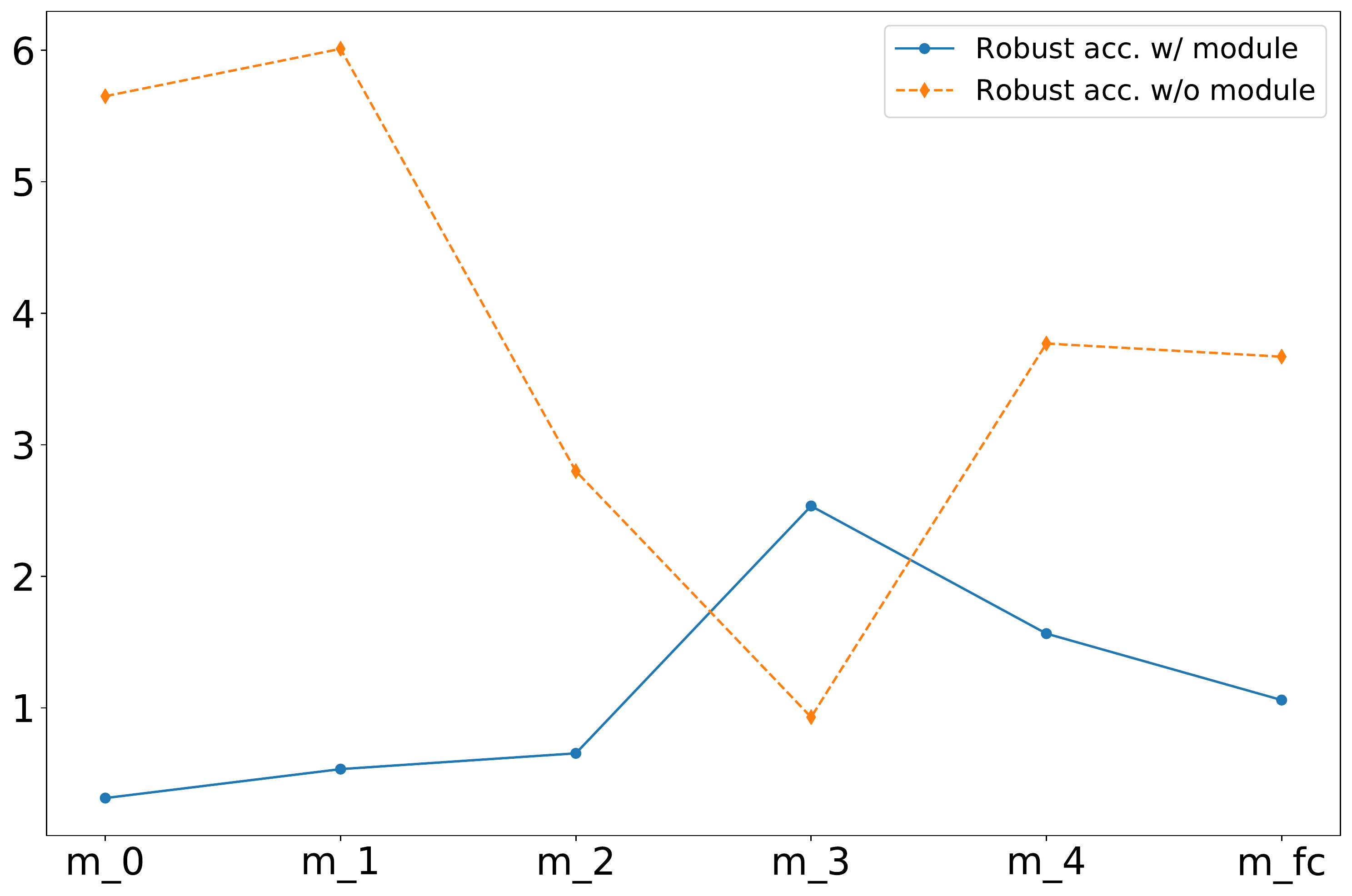}
        \caption{Conventional retraining}
    \end{subfigure}
    
    \caption{Median accuracy when considering the distribution of accuracies (ResNet-18 on CIFAR-10) when either including or excluding a particular module. Median robust performance drops whenever the initial layers are not adversarially trained in the end.}
    \label{fig:cifar10_resnet18_module_results}
\end{figure}

In our first set of experiments, we considered a single split in the network where we either train the network before the cut-off or after the cut-off while keeping the other part fixed.
However, this does not preclude the possibility that a combination of lower-level and higher-level modules provides better robustness as compared to just a single split.
In order to test this, we trained ResNet-18 (CIFAR-10) on all the different possible combinations of modules.
These results are summarized in Fig.~\ref{fig:cifar10_resnet18_module_results} where we plot the median accuracy for all possible combinations of the modules with or without a particular module.
These results are qualitatively consistent with single cut-off experiments, where adversarial training of the initial modules is essential for robustness to adversarial samples.
This indicates that there are no specific combinations of lower-level and high-level modules that are robust, but rather, just the initial set of layers are important for this purpose.

\subsection{Layer-wise Reinitialization Robustness}

\begin{figure*}[t]
\centering
    \begin{subfigure}[b]{\otherfigsize\textwidth}
        \includegraphics[width=\textwidth]{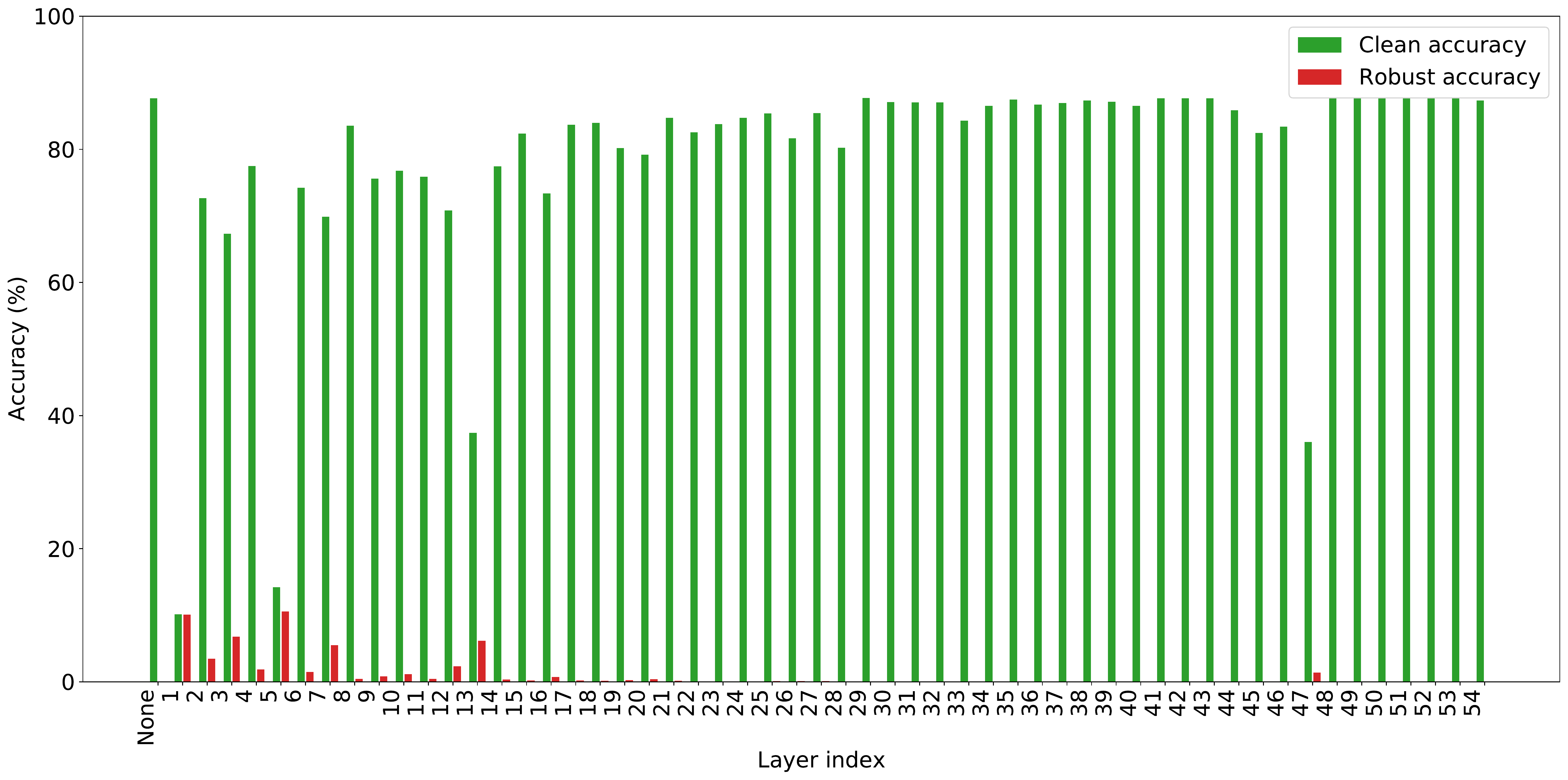}
        \caption{Conventionally pretrained (CIFAR-10)}
        \label{fig:reinit_con_cifar10}
    \end{subfigure}
    ~
    \begin{subfigure}[b]{\otherfigsize\textwidth}
        \includegraphics[width=\textwidth]{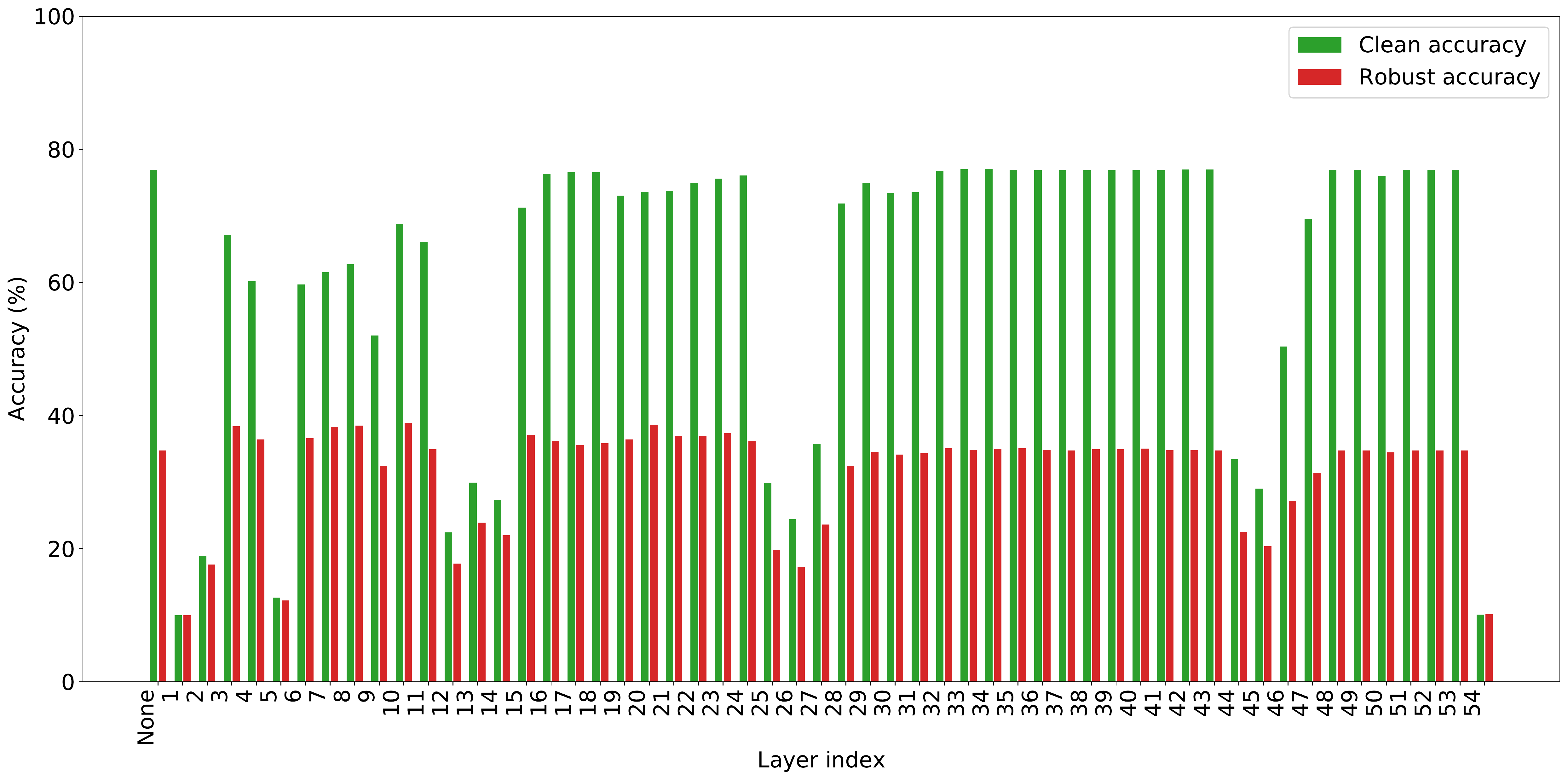}
        \caption{Adversarially pretrained (CIFAR-10)}
        \label{fig:reinit_adv_cifar10}
    \end{subfigure}
    
    \begin{subfigure}[b]{\otherfigsize\textwidth}
        \includegraphics[width=\textwidth]{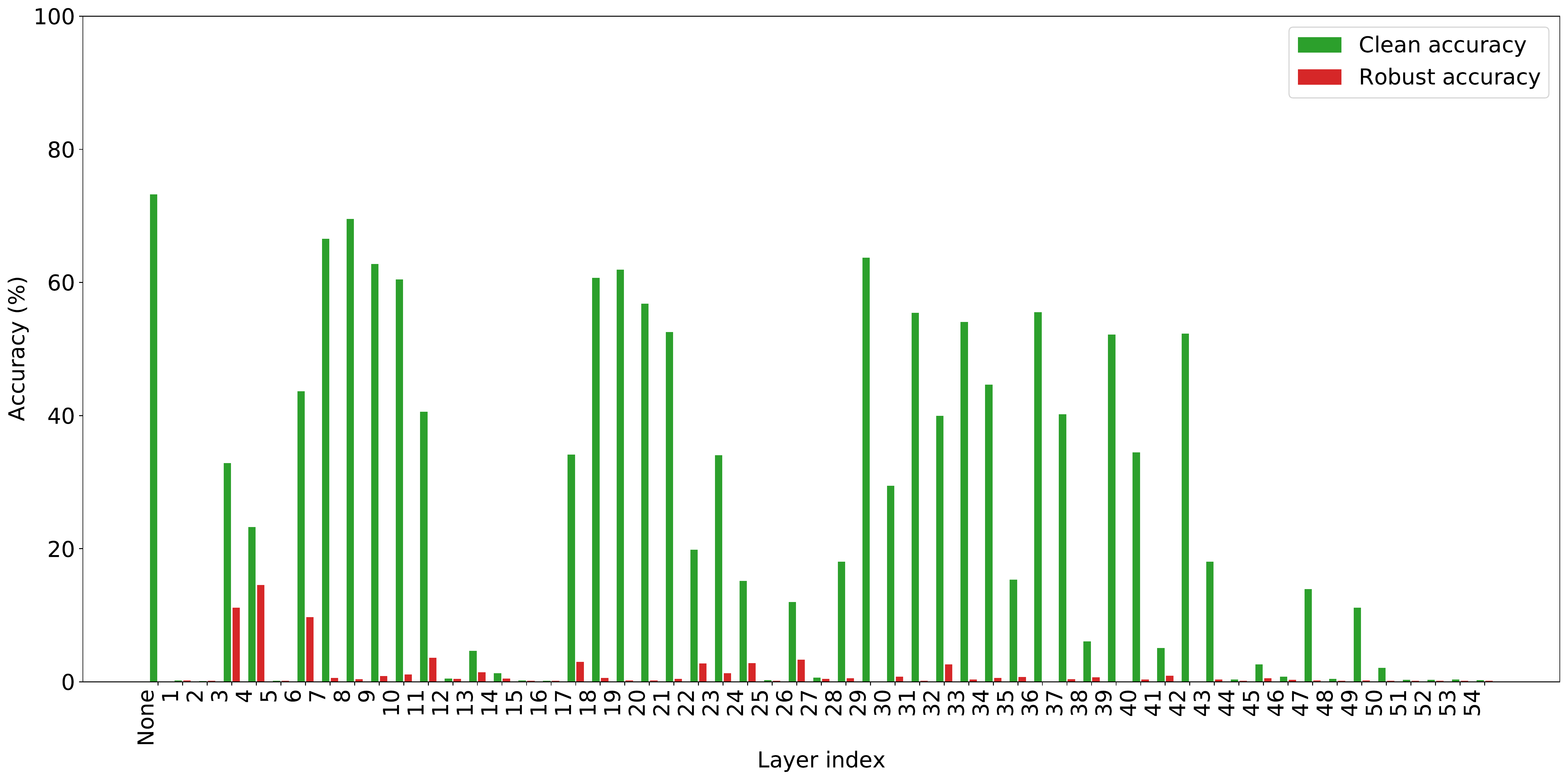}
        \caption{Conventionally pretrained (ImageNet)}
        \label{fig:reinit_con_imagenet}
    \end{subfigure}
    ~
    \begin{subfigure}[b]{\otherfigsize\textwidth}
        \includegraphics[width=\textwidth]{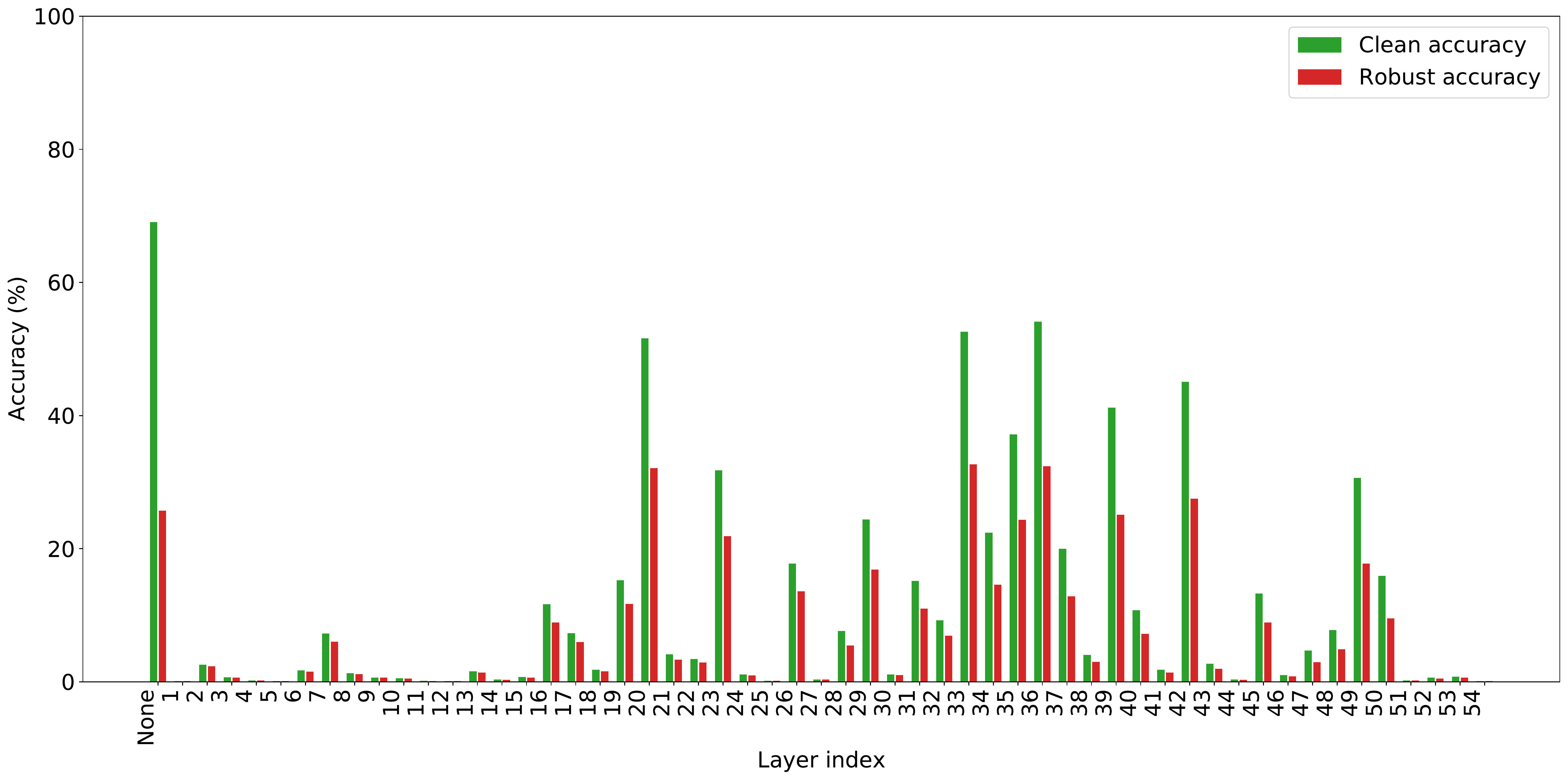}
        \caption{Adversarially pretrained (ImageNet)}
        \label{fig:reinit_adv_imagenet}
    \end{subfigure}
    
    \caption{Layer-wise reinitialization robustness for ResNet-50 on CIFAR and ImageNet with either conventional or adversarial training. In the figure, \textit{none} refers to the performance of the original model, while \textit{layer index} refers to the index of the convolutional layer in TorchVision's ResNet-50~\cite{torchvision}. Adversarially pretrained ResNet-50 demonstrates the high sensitivity of the initial layers in contrast to the conventionally pretrained model, which indicates that the initial layers change significantly to cater for the adversarial noise, highlighting their importance in obtaining robust models.}
    \label{fig:layer_sensitivity_results}
\end{figure*}

We attempt to understand the behavior described by Zhang et al. (2019)~\cite{zhang2019layersensitivity} for adversarial samples, where they found some layers to be much more important than others for overall classification performance.
Although we reproduce their results on clean samples, from the point of view of adversarial training, we find (Fig.~\ref{fig:reinit_con_cifar10}) that reinitialization of low-level layers tends to induce significant adversarial robustness. This is consistent with our other findings, namely that adversarial samples are a phenomenon associated with low-level layers. In addition, it suggests that susceptibility to adversarial samples is associated with training~\cite{ilyas2019adversarialexamplesarenotbugs}. Conversely, susceptibility to adversarial samples is never significantly increased due to layer reinitialization (Fig.~\ref{fig:reinit_adv_cifar10}).  Since ResNet-50 models trained on full ImageNet are much more susceptible to layer reinitialization, the results are more difficult to interpret. However, we still observe that reinitialization of initial layers tends to induce higher robustness to adversarial samples (Fig.~\ref{fig:reinit_con_imagenet}).
Furthermore, while non-adversarial accuracy is usually strongly affected by reinitialization, adversarial accuracy is usually less affected in comparison (Fig.~\ref{fig:reinit_adv_imagenet}).

\section{Feature Distributions}

Above, we have seen the differential effects of early and late layers on adversarial robustness.
Adversarial attacks on a network might operate by changing one type of feature into another, leaving the overall distribution of feature vectors the same, or by producing novel feature vectors that do not occur in non-adversarial samples.
These changes might occur only in late layers or both in early and late layers.
This distinction is important both for understanding the nature of adversarial attacks and to devise possible defenses.

Prior work visualizing the activities in hidden convolutional layers has primarily focused on visualizing the aggregate or per-filter activity~\cite{rauber2016visualizing}. In contrast, we visualize the distribution of activations for all the filters simultaneously across many images and layers, under both adversarial and non-adversarial conditions, using nonlinear dimensionality reduction by picking a single vector across spatial dimensions of the activation ($\mathbf{z} \in \mathbb{R}^{C}$ where $C$ is the number of filters in a layer).
Representative results are shown in Figure 8 using t-SNE dimensionality reduction. Images represent random samples after blocks one through four in a ResNet 50 network, choosing 100 random samples from each of 1000 different images. 
These results are consistent across three dimensionality reduction techniques (UMAP, TriMap, t-SNE) as well as two different datasets which we tested (ImageNet as well as CIFAR-10).
Each scatterplot is overlayed with a kernel density estimation in the dimensionality-reduced space, with green regions corresponding to non-adversarial samples and red regions corresponding to adversarial samples.

\begin{figure*}[t]
\centering
    \begin{subfigure}[b]{1.0\textwidth}
        \includegraphics[width=\textwidth]{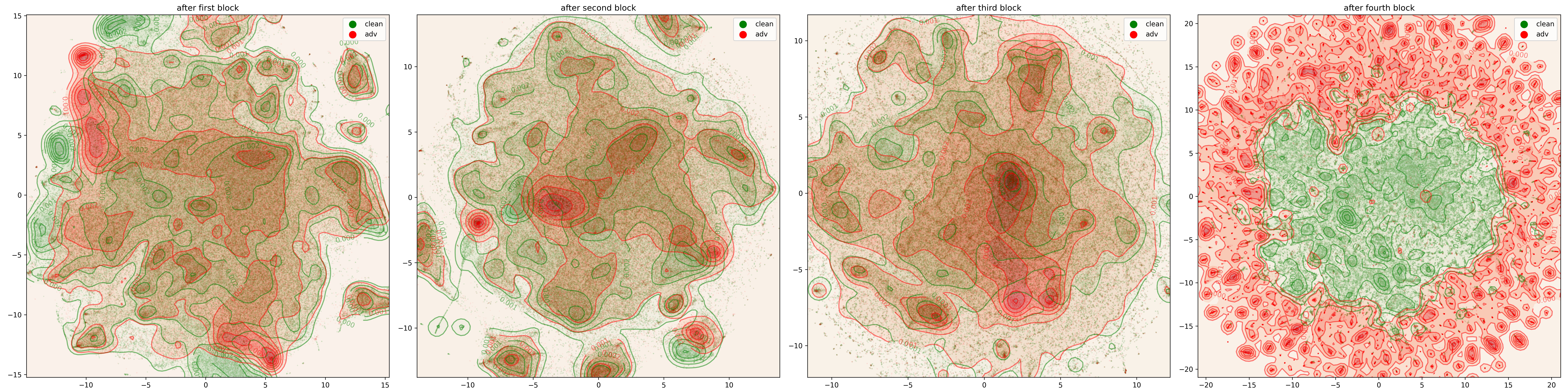}
        \caption{Conventionally pretrained ResNet-50 (ImageNet)}
        \label{fig:act_con_imagenet}
    \end{subfigure}
    
    \begin{subfigure}[b]{1.0\textwidth}
        \includegraphics[width=\textwidth]{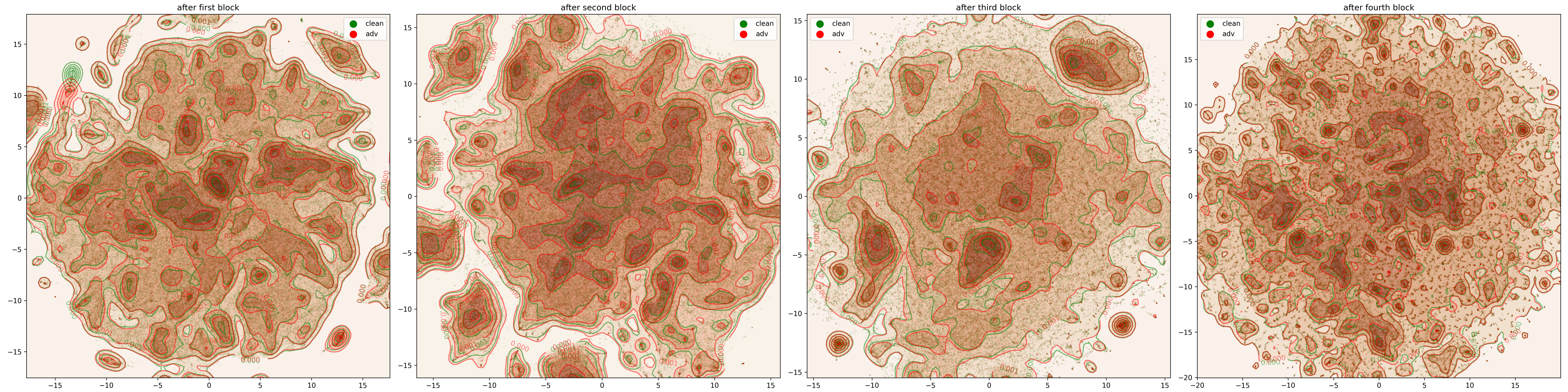}
        \caption{Adversarially pretrained ResNet-50 (ImageNet)}
        \label{fig:act_adv_imagenet}
    \end{subfigure}
    
    \caption{Dimensionality reduced activation vectors using t-SNE~\cite{tsne} of four (\textit{m\_1} to \textit{m\_4}) modules in the network. These plots highlight the substantial differences between clean and adversarial activations, which are amplified upon propagation through higher-level modules. Adversarial training minimizes these differences between activations.}
    \label{fig:activation_vec_dim_red}
\end{figure*}

Fig.~\ref{fig:act_con_imagenet} shows that the distribution of feature vectors differs substantially between non-adversarial and adversarial samples. In fact, after block 4 (high-level features), the distribution of adversarial and non-adversarial sample are almost completely non-overlapping, showing that adversarial samples do not imitate non-adversarial activation patterns or outputs, but generate very different collections of high-level features. This difference is particularly striking since there is little overlap even in the feature vectors that might correspond to background regions in the image. The second striking phenomenon observable in Fig.~\ref{fig:act_con_imagenet} is that substantial distributional differences are present even after the first block, i.e., in low-level features. This is consistent with our findings above, namely that differences between non-adversarial and adversarial samples must occur already in the early layers of the network and are responsible for susceptibility to adversarial samples.

Fig.~\ref{fig:act_adv_imagenet} illustrates the same distributions for adversarially trained networks. What we find here is that the distributions of feature vectors associated with non-adversarial and adversarial samples are much closer, not just overlapping in general, but reproducing peaks and regions of high density in substantial detail. That is, adversarial training has to adjust not just the high-level convolutional layers to match distributions between adversarial and non-adversarial samples, but also the low-level feature extraction layers (and from our above experiments, we already know that this change to the distribution of low-level feature vectors is both necessary and sufficient).

State-of-the-art reductions in susceptibility to adversarial samples through adversarial training results in feature vector distributions for non-adversarial and adversarial samples that closely match each other. Nevertheless, the resulting models still have substantial susceptibility to adversarial samples.
This implies that the remaining successful adversarial attacks probably work by transforming feature vectors into each other while staying within the distribution of non-adversarial feature vectors.
In other words, adversarial attacks on adversarially trained networks are qualitatively different from adversarial attacks on undefended networks.

\section{Conclusion}

We have described selective retraining of networks as a technique for localizing susceptibility to adversarial samples in deep neural networks.
Furthermore, we have demonstrated that dimensionality reduction of sets of activation vectors in different layers can be a useful tool for understanding the statistics and relations of adversarial and non-adversarial vectors.
Our experimental results demonstrate that susceptibility of deep neural networks to adversarial samples is associated with the early, non-specific layers of such networks.
That is, we have shown that adversarial samples generate differences in feature distributions in those layers and that training networks to be robust to adversarial samples largely eliminates those distributional differences.
Practically, this means that in order to achieve robustness to adversarial samples, it is both necessary and sufficient to retrain only the early layers where feature vectors are not yet highly class specific.
Our experiments also show that adversarial samples can be detected and visualized easily as
anomalies or outliers.
However, a substantial gap between human performance and deep neural networks however remains even after adversarial training. Our results show that these differences are not merely quantitative in nature; rather, in the absence of adversarial training, adversarial samples succeed via generating novel feature vectors, while after adversarial training, adversarial samples mimic the feature distribution of non-adversarial samples, suggesting that different defense mechanisms may be required. 

Our work has a number of practical implications: (1) it shows that we cannot quickly transform non-robust networks into robust networks by retraining, (2) adversarially trained networks are still susceptible to some adversarial attacks, but our results show that the methods used for detecting attacks on undefended networks fail for adversarially trained networks, and (3) we can likely achieve better detection of adversarial samples by analyzing unit outputs as a distribution of feature vectors rather than as a single vector.
The techniques described should prove useful in future work on understanding the statistical origins of adversarial samples, as well as devising practical techniques for defending against adversarial samples.

\section{Broader Impact}

Our investigation aims to help understand the causes of the existence of adversarial examples, which is a major failure mode of current deep learning models.
Deep learning-based visual recognition systems have been deployed in a range of different areas, including self-driving cars and security systems.
Improving the robustness of these systems is critical. 
Furthermore, a better understanding of robustness can help us achieve robustness in an efficient way without going through the compute-intensive process of adversarial training.
However, on the flip side, these robust systems can potentially be used in a negative context such as mass surveillance.

\section*{Acknowledgements}

The authors would like to acknowledge useful discussions with Iuri Frosio on adversarial robustness.
This work is in part supported by the BMBF project DeFuseNN (Grant 01IW17002) and the NVIDIA AI Lab (NVAIL) program.

\bibliographystyle{plain}
\bibliography{bibliography.bib}

\end{document}